%% file: main.tex
\documentclass{article}

\usepackage[final,nonatbib]{neurips_data_2021}
\usepackage[numbers]{natbib}

\usepackage[utf8]{inputenc} % allow utf-8 input
\usepackage[T1]{fontenc}    % use 8-bit T1 fonts
\usepackage{hyperref}       % hyperlinks
\usepackage{url}            % simple URL typesetting
\usepackage{booktabs}       % professional-quality tables
\usepackage{amsfonts}       % blackboard math symbols
\usepackage{nicefrac}       % compact symbols for 1/2, etc.
\usepackage{microtype}      % microtypography
\usepackage{xcolor}         % colors

\usepackage{amsmath,amsthm,amssymb}
\usepackage{algorithm}
\usepackage{algorithmic}
\usepackage{comment}
\usepackage{mathtools}
\usepackage{caption}
\usepackage{float}
\usepackage{graphicx}
\usepackage{subfigure}
\usepackage{multirow}
\usepackage{mathrsfs}
\usepackage{newfloat}
\usepackage{relsize}
\usepackage{enumitem}
\usepackage{nicefrac}
\usepackage{pgfplots}
\usepackage{graphicx}
\usepackage{bm}

\usepackage{pifont}

\newcommand{\D}{\mathcal{D}}
\newcommand{\Dtr}{\mathcal{D}_{\text{train}}}
\newcommand{\Dte}{\mathcal{D}_{\text{test}}}
\newcommand{\E}{\mathbb{E}}
\newcommand{\ind}{\mathbb{I}}
\newcommand{\reponame}{\textsc{XAI-Bench}}

\title{Synthetic Benchmarks for Scientific Research in Explainable Machine Learning}

\author{%
  Yang Liu\thanks{Equal contribution. 
  } \\
  Abacus.AI\\
  San Francisco, CA 94103 \\
  \texttt{yang@abacus.ai} \\
  \And
  Sujay Khandagale\footnotemark[1] \\
  Abacus.AI\\
  San Francisco, CA 94103 \\
  \texttt{sujay@abacus.ai} \\
  \And
  Colin White \\
  Abacus.AI\\
  San Francisco, CA 94103 \\
  \texttt{colin@abacus.ai} \\
  \And
  Willie Neiswanger \\
  Stanford University\\
  Stanford, CA 94305 \\
  \texttt{neiswanger@cs.stanford.edu} \\
}

\begin{document}

\maketitle

\input{sections/1-abstract}
\input{sections/2-introduction}
\input{sections/3-related-work}

\input{sections/4-synthetic-datasets}
\input{sections/5-experiments}

\input{sections/6-impact}
\input{sections/10-conclusion}

\begin{ack}
Work done while the first three authors were working at Abacus.AI.
WN was supported by U.S. Department of Energy Office of Science under Contract No. DE-AC02-76SF00515.
\end{ack}

%%%%%%%%%%%%%%%%%%%%%%%%%%%%%%%%%%%%%%%%%%%%%%%%%%%%%%%%%%%%
\newpage

\bibliography{main}
\bibliographystyle{plain}

\newpage

\newpage
\appendix

\input{sections/99-appendix}

\end{document}

%% file: sections/1-abstract.tex
\begin{abstract}
\vspace{-3mm}

As machine learning models grow more complex and their applications become more high-stakes,
tools for explaining model predictions have become increasingly important.
This has spurred a flurry of research in model explainability and has given rise to feature attribution
methods such as LIME and SHAP.
Despite their widespread use, evaluating and comparing different feature attribution 
methods remains challenging: evaluations ideally require human studies, and empirical evaluation metrics 
are often data-intensive or computationally prohibitive on real-world datasets.
In this work, we address this issue by releasing 
\reponame: a suite of synthetic datasets along with a library
for benchmarking feature attribution algorithms. Unlike real-world datasets, synthetic datasets allow 
the efficient computation of conditional expected values that are needed to evaluate ground-truth
Shapley values and other metrics.
The synthetic datasets we release offer a wide variety
of parameters that can be configured to simulate real-world data.
We demonstrate the power of our library by benchmarking popular explainability techniques across several evaluation metrics and across a variety of settings.
The versatility and efficiency of our library will help researchers bring their explainability methods 
from development to deployment.
Our code is available at \url{https://github.com/abacusai/xai-bench}.
\end{abstract}

\begin{comment}

% inline

As machine learning models grow more complex and their applications become more high-stakes, tools for explaining model predictions have become increasingly important. This has spurred a flurry of research in model explainability and has given rise to feature attribution methods such as LIME and SHAP. Despite their widespread use, evaluating and comparing different feature attribution methods remains challenging: evaluations ideally require human studies, and empirical evaluation metrics are often data-intensive or computationally prohibitive on real-world datasets. In this work, we address this issue by releasing XAI-Bench: a suite of synthetic datasets along with a library for benchmarking feature attribution algorithms. Unlike real-world datasets, synthetic datasets allow the efficient computation of conditional expected values that are needed to evaluate ground-truth Shapley values and other metrics. The synthetic datasets we release offer a wide variety of parameters that can be configured to simulate real-world data. We demonstrate the power of our library by benchmarking popular explainability techniques across several evaluation metrics and across a variety of settings. The versatility and efficiency of our library will help researchers bring their explainability methods from development to deployment. Our code is available at https://github.com/abacusai/xai-bench.

\end{comment}

%% file: sections/2-introduction.tex
\vspace{-3mm}
\section{Introduction}\label{sec:introduction}
\vspace{-2mm}

The last decade has seen a rapid increase in applications 
of machine learning in a wide variety of high-stakes domains, such as
credit scoring, fraud detection, criminal recidivism, and
loan repayment~\cite{mukerjee2002multi, bogen2018help, ngai2011application, barocas2017fairness}.
With the widespread deployment of machine learning models in applications that impact human
lives, research on model explainability has become increasingly important.
The applications of model explainability include debugging, legal obligations to give explanations, recognizing and mitigating bias, 
data labeling, and faster adoption of machine learning 
technologies~\citep{shap, zhang2020survey, arya2019one, deyoung2019eraser}.
Many different methods for explainability are actively being explored, 
including logic rules~\citep{fu1991rule, towell1993extracting, setiono1995understanding},
hidden semantics~\citep{zhang2018interpretable},
feature attribution~\citep{lime, shap, maple, l2x, strumbelj2010efficient},
and explanation by example~\citep{li2018deep, chen2018looks}.
The most common type of explainers are post-hoc, local feature attribution
methods~\citep{zhang2020survey, shap, shapr, lime, maple, l2x}, 
which output a set of weights corresponding to the importance of each feature for a given
datapoint and model prediction.
Although various feature attribution methods are being deployed in different use cases
today, currently there are no widely adopted methods to easily
\emph{evaluate and/or compare} different feature attribution algorithms.
Indeed, evaluating the effectiveness of explanations is an intrinsically 
human-centric task that ideally requires human studies.
However, it is often desirable to develop new explainability techniques using empirical evaluation metrics before the human trial stage.
Although empirical evaluation metrics have been proposed, 
many of these metrics are either computationally prohibitive
or require strong assumptions, to compute on real-world datasets.
For example, a popular method for feature attribution is to approximate Shapley
values~\citep{shap, datta2016algorithmic, lipovetsky2001analysis, strumbelj2010efficient}, 
but computing the distance to ground-truth Shapley values 
requires estimating
exponentially many conditional feature distributions, which is 
not possible to compute unless the dataset contains sufficiently many datapoints 
across exponentially many combinations of features.

%%% could also mention new metrics

In this work, we overcome these challenges by releasing a suite of synthetic datasets, which make it
possible to efficiently benchmark feature attribution methods.
The use of synthetic datasets, for which the ground-truth distribution of data is known, makes it
possible to exactly compute the conditional distribution over any set of features, thus enabling
computations of many feature attribution evaluation metrics such as
distance to ground-truth Shapley values~\citep{shap}, 
remove-and-retrain (ROAR)~\citep{roar}, 
faithfulness~\citep{alvarez2018towards}, monotonicity~\citep{luss2019generating},
and infidelity~\citep{yeh2019fidelity}.
Our synthetic datasets offer a wide variety of parameters which can be configured to simulate
real-world data and have the potential to identify subtle failures,
such as the deterioration of performance on datasets with high feature correlation.
We give examples of how real datasets can be converted to
similar synthetic datasets, thereby allowing
explainability methods to be benchmarked on realistic synthetic datasets.

We showcase the power of our library by benchmarking popular explainers such as 
SHAP~\citep{shap}, LIME~\citep{lime}, MAPLE~\citep{maple}, SHAPR~\citep{shapr}, 
L2X\citep{l2x}, and breakDown~\citep{breakdown},
on a broad set of evaluation metrics, across a variety of 
axes of comparison, such as feature correlation, model type, and data distribution type.
Our library is designed to substantially reduce the time required for researchers
and practitioners to move their explainability algorithms from development to deployment.
Our code, API docs, and raw experimental results are available at \url{https://github.com/abacusai/xai-bench}.
%, and we give a resubmission statement in Appendix~\ref{app:statement}.
We welcome contributions and hope to grow the repository to handle a wide variety of use-cases.

\noindent\textbf{Our contributions.}
We summarize our main contributions below.
\begin{itemize}[topsep=0pt, itemsep=2pt, parsep=0pt, leftmargin=5mm]
    \item 
    We release a set of synthetic datasets with known ground-truth distributions,
    along with a library that makes it possible to efficiently evaluate feature attribution
    techniques with respect to popular evaluation metrics. 
    Our synthetic datasets offer a number of parameters that can be configured to simulate real-world applications.
    \item 
    We demonstrate the power of our library by benchmarking popular explainers such as
    SHAP~\citep{shap}, LIME~\citep{lime}, MAPLE~\citep{maple}, SHAPR~\citep{shapr}, L2X\citep{l2x},
    and breakDown~\citep{breakdown}.
\end{itemize}

\input{figures/overview}

%% file: figures/overview.tex
%% OVERVIEW FIGURE
\begin{figure}[t!]
    \centering
    \includegraphics[width=0.9\linewidth]{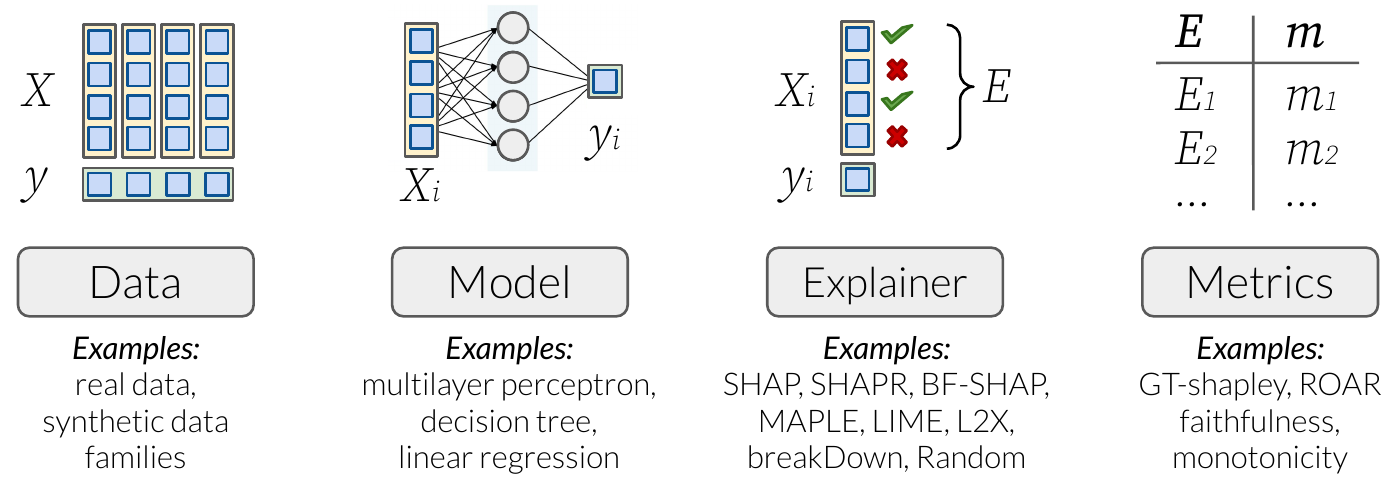}
    \vspace{-1mm}
    \caption{Overview of the main components in \reponame.}
    \label{fig:overview}
    \vspace{-4mm}
\end{figure}

%% file: sections/3-related-work.tex
\vspace{-2mm}
\section{Related Work}\label{sec:relatedwork}
\vspace{-1mm}

%%% overview

Model explainability in machine learning has seen a wide range of approaches,
and multiple taxonomies have been proposed to classify the different types of approaches.
Zhang et al.~\citep{zhang2020survey} describe three dimensions of explainability techniques:
passive/active, type of explanation, and local/global explainations.
The types of explanations they identified are 
logic rules~\citep{fu1991rule, towell1993extracting, setiono1995understanding},
hidden semantics~\citep{zhang2018interpretable},
feature attribution~\citep{lime, shap, maple, l2x, strumbelj2010efficient, shapr},
and explanation by example~\citep{li2018deep, chen2018looks}.
Other surveys on explainable AI include Arrieta et 
al.~\citep{arrieta2020explainable}, Adadi and Berrada~\citep{adadi2018peeking},
and Do{\v{s}}ilovi{\'c} et al.~\citep{dovsilovic2018explainable}.

%%% feature attribution
Techniques for feature attribution include approximating Shapley 
values~\citep{shap, datta2016algorithmic, lipovetsky2001analysis, strumbelj2010efficient},
approximating the model locally with a more explainable model~\citep{lime},
and approximating the mutual information of each feature with 
the label~\citep{l2x}.
%In Appendix~\ref{app:descriptions}, we give descriptions and implementation details of the feature attribution methods we implemented.
%%%% failure modes
Other work has also identified failure modes for some explanation techniques.
For example, recent work has shown that explanation techniques are susceptible to adversarial feature 
perturbations~\citep{dombrowski2019explanations, slack2020fooling, heo2019fooling},
high feature correlations~\citep{lakkaraju2020fool},
and small changes in hyperparameters~\citep{garreau2020explaining, bansal2020sam}.

\begin{comment}
% failure mode notes:
\citep{dombrowski2019explanations}
Show that explanations can be manipulated arbitrarily by applying hardly perceptible perturbations
Only try it on gradient-based methods
\citep{slack2020fooling}
They fool LIME and SHAP using adversarial attacks
\citep{lakkaraju2020fool}
They show how explanations can be misleading due to correlations in the features
They rely on user studies
\citep{heo2019fooling}
They fool Grad-CAM, LRP, and SimpleGrad with adversarial examples

Some work constructs adversarial examples (mostly focusing on computer vision datasets). Another work shows how explanations can be misleading due to correlations, using a theoretical analysis and user study. 
Our work is the first to quantify the relationship between empirical evaluations of explainers with respect to properties of the dataset such as feature correlation.

\end{comment}

%%% experimental survey / benchmark-type papers
\vspace{-1mm}
\subsection{Benchmarking Explainability Techniques}
\vspace{-1mm}
One recent work~\citep{jeyakumar2020can} gave an experimental survey of explainability methods,
testing SHAP~\citep{shap}, LIME~\citep{lime}, Anchors~\citep{ribeiro2018anchors}, 
Saliency Maps~\citep{simonyan2013deep}, Grad-CAM++~\citep{chattopadhay2018grad}, 
and their proposed~ExMatchina on image, text, audio, and sensory datasets.
They use human labeling via Mechanical Turk as an evaluation metric.
Another work~\citep{arya2019one} gave an experimental survey of several algorithms
including local/global, white-box/black-box, and supervised/unsupervised techniques.
The only feature attribution algorithms they tested were SHAP and LIME.
Other recent work gives a benchmark on explainability for time-series 
classification~\citep{fauvel2020performance}, 
or for natural language processing (NLP)~\citep{deyoung2019eraser}.
%datasets aimed at comparing explainability methods. 
%This work releases multiple datsets with human-annotated explanations, as well as a few newly proposed metrics specifically chosen to capture the explainability of predictions in NLP applications.
%
Finally, concurrent work~\citep{leaf} releases a library with several evaluation metrics
for local linear explanation methods and uses the library to compare LIME and SHAP.
To the best of our knowledge, no prior work has released a library with five 
different evaluation metrics or released a set of synthetic datasets for explainability 
with more than one tunable parameter.

\vspace{-1mm}
\subsection{Metrics}
\vspace{-1mm}

While the ``correctness'' of feature attribution methods may be subjective
or application-specific~\citep{yang2019benchmarking},
comparisons between methods are often based on 
human studies~\citep{lage2018human, ross2018improving, selvaraju2017grad}.
However, human studies are not always possible, and several empirical (non-human)
evaluation metrics have been proposed.
Faithfulness~\citep{alvarez2018towards, arya2019one, adebayo2018sanity, dhurandhar2019model, lakkaraju2019faithful}, 
infidelity~\citep{yeh2019fidelity, bhatt2020evaluating, samek2016evaluating}, and
monotonicity~\citep{luss2019generating, arya2019one, das2020opportunities}
are popular explainability metrics
which measure whether each feature's susceptibility to change the model output
is aligned with each feature's attribution weight.
Another popular metric, remove-and-retrain 
(ROAR)~\citep{roar, hartley2020explaining, hartley2021swag, meng2021mimic}, 
measures these statistics by retraining the model each time relevant features are removed,
in order to avoid inaccuracies due to distribution shift.
In the next section, we give the formal definition and a discussion 
for each metric.

%  By retraining a model with subsets of features ablated, ROAR

\begin{comment}

Faithfulness~\citep{alvarez2018towards} measures the correlation between the weights of the
feature attribution algorithm, and the effect of the features on the performance of the model.
%
Monotonicity~\citep{luss2019generating} checks whether iteratively adding features from
least weighted feature to most weighted feature, causes the prediction to monotonically improve.
%
By retraining a model with subsets of features ablated, ROAR~\citep{roar} uses a new model with partially ablated input features to evaluate a feature attribution technique while avoiding
problems with distribution shift.
%
Note that all of the above metrics evaluate feature importance by computing the effect of 
removing the feature from a single set of features $S$. 
In contrast, Shapley 
values~\citep{shap, datta2016algorithmic, lipovetsky2001analysis, strumbelj2010efficient} 
evaluate all possible sets $S$ that a feature can be removed from 
to compute an average effect.
\end{comment}

%% file: sections/4-synthetic-datasets.tex
\vspace{-2mm}
\section{Evaluation Metrics}\label{sec:evaluationmetrics}
\vspace{-1mm}

\subsection{Preliminaries}
\label{sec:prelim}
\vspace{-1mm}
We first give definitions and background information used throughout the next three
sections.
Given a distribution $\D$,
each datapoint is of the form $(\bm{x}, y)\sim \D$, where $\bm{x}$ denotes the set of
features, and $y$ denotes the label. We assume that $\bm{x}\in [0,1]^D$, yet all of the concepts we discuss can be generalized to arbitrary categorical
and real-valued feature distributions.
Assume we have a training set $\Dtr$ and a test set $\Dte$, 
both drawn from $\D$.
For the case of regression, we train a model $f:[0,1]^D\rightarrow [0,1]$ on the training set. We also implement classification using cross-entropy loss.
Common choices for $f$ include a neural network or a decision tree.

A \emph{feature attribution method} is a function $g$ which can be used to estimate the importance
of each feature in making a prediction. That is, given a model $f$ and a datapoint $\bm{x}$, then
$g(\bm{x}, f)=\bm{w}\in \mathcal{R}^D$,
where each output weight $w_i$ corresponds to the relative
importance of feature $i$ when making the prediction $f(\bm{x})$.
Common choices for $g$ include SHAP~\citep{shap} or LIME~\citep{lime}.
%We discuss feature attribution methods in Section~\ref{app:experiments}.

\vspace{-1mm}
\subsection{Metrics}\label{sec:metrics} 
\vspace{-1mm}

In this section, we formally define popular evaluation metrics for explainability 
methods. Each evaluation metric has pros and cons and may be more or less appropriate
depending on the application and problem instance.
We provide a guide to choosing metrics in Section~\ref{subsec:guide}.

A \emph{feature attribution evaluation metric} is a function
that evaluates the weights of a feature attribution method on a datapoint $\bm{x}$.
For example, given a datapoint $\bm{x}$ and a set of feature weights $\bm{w}=g(\bm{x}, f)$,
then a value near or below zero indicates that $g$ did not provide an accurate
feature attribution estimate for $\bm{x}$, while a value near one indicates
that $g$ did provide an accurate feature attribution estimate.

Many evaluation metrics involve evaluating the change in performance of the model
when a subset of features of a datapoint are removed. 
In order to measure the true marginal improvement for a set of features $S$, one approach is
to evaluate the model when replacing the features $S$ with their expected values conditioned on 
the remaining features~\citep{datta2016algorithmic, sundararajan2020many, chen2020true, janzing2020feature}.
Formally, given a datapoint $\bm{x}\sim \D$ and a set of indices 
$S \subseteq \{1,\cdots,D\}$, we define
$\D\left(\bm{x}_{S}\right)$ as the conditional probability distribution $\bm{x}'\sim\D$ such 
that $x'_i=x_i$ for all $i\in S$. In other words, given $\bm{x}$ and $S$, we have
\begin{equation}\label{eq:conditional}
    p\left(\bm{x}'\sim\D\left(\bm{x}_{S}\right)\right) = p\left(\bm{x}'\sim\D\mid x_i'=x_i\text{ for all }i\in S\right).
\end{equation}
By this definition, $\D\left(\bm{x}_{\emptyset}\right)=\D$, and if we define
$F=\{1,\cdots,D\}$, then $\bm{x}'\sim\D\left(\bm{x}_{F}\right)$ is equal to
$\bm{x}$ with probability 1.
Later in this section, we discuss other popular choices such as interventional
conditional distributions~\citep{lundberg2017unified, shapr}.
Given a datapoint $\bm{x}$, a model $f$, and a weight vector $\bm{w}$,
the first evaluation metric, \textbf{faithfulness}~\citep{alvarez2018towards},
is defined as follows:
\begin{equation}
    \text{faithfulness} = \text{Pearson}
        \left(\left|\E_{\bm{x}'\sim \D\left(\bm{x}_{F \setminus i}\right)}[f(\bm{x}')] -f(\bm{x})\right|_{1\leq i\leq D}, 
        [w_i]_{1\leq i\leq D}\right).
\end{equation}
Intuitively, faithfulness computes the Pearson correlation 
coefficient~\citep{wright1921correlation} between the weight vector $\bm{w}$ and
the approximate marginal contribution 
$\left|\E_{\bm{x}'\sim \D\left(\bm{x}_{F \setminus i}\right)}[f(\bm{x}')] -f(\bm{x})\right|$
for each feature $i$.
Faithfulness is a lightweight metric that is especially useful for comparing which
feature would have the most impact on the model output when individually changed.

The next metric computes the marginal improvement of each feature ordered by the
weight vector $\bm{w}$ \emph{without replacement}, and then computes the fraction of indices $i$
such that the marginal improvement for feature $i$ is greater than the marginal improvement for
feature $i+1$.
This makes it useful when comparing the effect of features as they are 
added sequentially.
Formally, define $S^+(\bm{w}, i)$ as the set of $i$ most important weights,
%define $S^-(\bm{w}, i)$ as the set of $i$ least important weights,
and let $S^+(\bm{w}, 0) = \emptyset$. 
Given a datapoint $\bm{x}$, a model $f$, and a weight vector $\bm{w}$,
\textbf{monotonicity}~\citep{luss2019generating} is defined as follows:
% \begin{align}
%     \delta_{i}^{+} \quad &= \quad
%     \E_{\bm{x}'\sim \D\left(\bm{x}_{S^{+}(\bm{w}, i+1)}\right)}[f(\bm{x}')]
%     -\E_{\bm{x}'\sim \D\left(\bm{x}_{S^{+}(\bm{w}, i)}\right)}[f(\bm{x}')],\\
%     \text{monotonicity} &= \quad
%     \frac{1}{D-1}\sum_{i=0}^{D-2}  \ind_{|\delta_{i}^{+}|\leq
%     |\delta_{i+1}^{+}|}.
% \end{align}
\begin{align}
    \text{monotonicity} \; &= \;
    \frac{1}{D-1}\sum_{i=0}^{D-2}  \ind_{|\delta_{i}^{+}|\leq
    |\delta_{i+1}^{+}|},\\
    \text{where} \hspace{2mm}
    \delta_{i}^{+} \; &= \;
    \E_{\bm{x}'\sim \D\left(\bm{x}_{S^{+}(\bm{w}, i+1)}\right)}[f(\bm{x}')]
    -\E_{\bm{x}'\sim \D\left(\bm{x}_{S^{+}(\bm{w}, i)}\right)}[f(\bm{x}')].
\end{align}

The types of metrics discussed so far all evaluate weight
vectors by comparing an estimate of the marginal improvement of a set of features to their
corresponding weights.
Estimating the marginal improvement requires computing $f$ on different combinations
of features, and it is possible that these combinations of features have very low density
in $\D$, and are therefore unlikely to occur in $\Dtr$.
This is especially true for structured data or data where there are large low-density
regions in $\D$ that may make the evaluations on $f$ unreliable. 
To help mitigate this issue, another paradigm of explainability evaluation metrics was
proposed: remove-and-retrain (\textbf{ROAR})~\citep{roar}. In this paradigm, in order to evaluate
the marginal improvement of sets of features, the model is retrained using a new dataset
with the features removed.
%
% todo check this
For example, rather than computing $|\E_{\bm{x}'\sim \D(\bm{x}_{F \setminus i})}[f(\bm{x}')] -f(\bm{x})|$,
we would compute
$|f^*(\E_{\bm{x}'\sim \D(\bm{x}_{F \setminus i})}[\bm{x}']) -f(\bm{x})|$,
where $f^*$ denotes a model that has been trained on a modification of $\Dtr$
where each datapoint has its $i$ features with highest weight removed.
%
% In the original work, the ROAR metric plots the
The original work plots the retrained model performance versus the number of features ablated~\citep{roar},
removing features in order of decreasing importance. Then feature attribution methods
are compared by inspecting the steepness of these plots.
Follow-up work has compressed the ROAR statistic into a scalar value by computing
the area-under-the-curve (AUC)~\citep{hartley2020explaining, meng2021mimic}.
We use this AUC version in Section~\ref{sec:experiments}, to be consistent with the other
metrics that only output a single value.
Note that to compute ROAR on all datapoints in the test set, 
the explainer must evaluate all datapoints in the 
training set to construct $D+1$ ablated datasets, 
and then the model must be retrained for each of these datasets.
We give the formal definition in Appendix~\ref{app:descriptions}.

A caveat for all of the aforementioned metrics is that they evaluate each feature weight 
by computing the effect of removing the feature from a single set of features $S$. 
While this evaluation is sufficient in many cases,
it may lead to unreliable measurements for e.g.\ highly nonlinear models.
Furthermore, the explicit goal of a popular line of explainability methods is to
obtain fast and accurate approximations of 
\emph{Shapley values}~\citep{shap, shapr, lundberg2017consistent, datta2016algorithmic, lipovetsky2001analysis, strumbelj2010efficient}.
To address this, we consider a metric based on Shapley values, \textbf{GT-Shapley},
which computes the Pearson correlation coefficient~\citep{wright1921correlation} of the 
feature weights to the ground-truth Shapley values.
Shapley values take into account the marginal improvement of a feature $i$ across 
\emph{all possible} exponentially many sets with and without $i$.
%We give the formal definition in Appendix~\ref{app:descriptions}.

Next, we consider the \textbf{infidelity} metric~\citep{yeh2019fidelity}.
This metric is computed by considering the effects of replacing each feature with
a \emph{noisy} baseline conditional expectation. Instead of computing the
correlation between the feature importances and the change in function values
(as in faithfulness and GT-Shapley), infidelity computes the difference between 
the change in function value and the dot product of the change in feature value 
with the feature importance vector, in expectation over the noise.
Note that if we were to only add noise to one feature at a time, this would be similar in
spirit to faithfulness (since the dot product would be equal to the
weight of the feature which had noise added). 
Similar to prior work~\citep{yeh2019fidelity}, we consider perturbations based on 
Gaussian noise. Therefore, infidelity can pick up nonlinear trends in feature 
importances better than faithfulness or monotonicity.

% removed this table to have more space
%See Table~\ref{tab:metrics} for a summary of the properties of each metric.
% \input{tables/metrics}
%
Finally, while Equation~(\ref{eq:conditional}) defines ``observational'' conditional
expectations~\citep{shap, shapr}, we also implement ``interventional'' conditional 
expectations~\citep{datta2016algorithmic, sundararajan2020many},
which are defined by assuming the features in $S$ are independent of the 
remaining features. 
This can be applied to all metrics defined in this section.
The best choice of conditional expectations depends on the
application~\citep{chen2020true}, and we discuss the tradeoffs in the next section.

\vspace{-3mm}
\subsection{A guide to choosing metrics} \label{subsec:guide}
\vspace{-2mm}

All of the metrics listed above may be used for evaluating and comparing different feature attribution techniques.
However, each metric has strengths and weaknesses, 
and choosing the most useful metric for a given situation
depends on the use case, dataset, feature attribution technique, 
and computational constraints. We discuss strengths, weaknesses, and example
use cases of each metric type.

For the ROAR paradigm, retraining the model with the most
important features removed is especially important when the original model is not calibrated for
out-of-distribution predictions~\citep{roar}, such as in high-dimensional applications like 
computer vision~\citep{meng2021mimic, hartley2021swag, srinivas2019full}.
However, retraining might fail to give an accurate evaluation in the presence of high feature
correlations~\citep{nguyen2020quantitative}. Furthermore, retraining the model incurs
a much larger computational cost.%, which is especially important when a model is computationally intensive to train.
% Roar is useful for datasets with a high ratio of relevant features to datapoints (such as computer vision) when there are no computational constraints.

For some feature attribution algorithms, the explicit goal is to efficiently
approximate the Shapley values~\citep{shap, shapr, lundberg2017consistent, datta2016algorithmic, lipovetsky2001analysis, strumbelj2010efficient}, and
the GT-Shapley metric is the best choice to determine which technique gives the best 
approximations to the true Shapley values. However, evaluating the ground-truth Shapley 
values has a computational cost that is exponential in the number of features.
Therefore, the GT-Shapley metric is slow to evaluate on high-dimensional datasets.

Faithfulness, monotonicity, and infidelity are far less computationally intensive
compared to ROAR and GT-Shapley.
% 
% faithfulness vs monotonicity
The main difference between faithfulness and monotonicity is that faithfulness considers
subsets of features by iteratively removing the most important features 
\emph{with replacement}, while monotonicity does this \emph{without replacement}.
Therefore, the former is better for applications where the main question is which features
would individually change the output of the model on a given datapoint 
(and therefore may be better on datasets with less correlated features).
The latter is better for applications where the main goal is to see the cumulative effect
of adding features (and therefore performs comparatively better in the presence of 
correlated features).

The main difference between infidelity and faithfulness (as well as monotonicity)
is that infidelity considers ablations of subsets of features, while faithfulness
only considers ablating a single feature at a time. Therefore, infidelity may
be more appropriate for models with highly nonlinear feature interactions,
compared to faithfulness and monotonicity.

% interventional vs observational conditional expectations
Finally, we discuss using interventional versus observational conditional expectations.
As pointed out in prior work~\citep{chen2020true}, 
interventional conditional expectations are better for applications
that require being ``true to the model'', while observational conditional expectations are better for
applications that require being ``true to the data'', because observational conditional expectations
tend to spread out importance among correlated features (even features that are not used by the 
model).
For example, interventional conditional expectations are more appropriate in explaining why a model
caused a loan to be denied, while observational conditional expectations are more appropriate in
explaining the causal features in the drug response to RNA sequences~\citep{chen2020true}.
%

\begin{comment}

Now we conclude with an example of an application for each of the components of the metrics.

\begin{itemize}[topsep=0pt, itemsep=2pt, parsep=0pt, leftmargin=5mm]
    \item \textbf{ROAR} is useful for datasets with a high ratio of relevant features 
    to datapoints (such as computer vision) when there are no computational constraints.
    \item \textbf{Faithfulness} is useful on datasets with uncorrelated features, 
    such as low-dimensional tabular data.
    \item \textbf{Monotonicity} is useful on datasets with correlated features, 
    such as Census data that contains redundant information.
    \item \textbf{GT-Shapley} is useful for datasets with a small number of highly correlated
    features and especially when comparing Shapley-based explainers such as
    Kernel-SHAP, SHAPR, or other SHAP
    variants~\citep{strumbelj2010efficient, ancona2019explaining, heskes2020causal}.
    \item \textbf{interventional conditional expectations} are useful for causal
    tasks such as explaining the causal features in the drug response to RNA sequences.
    \item \textbf{observational conditional expectations} are useful for model-based tasks
    such as explaining the importance of each feature in a model that predicts loan repayment.
\end{itemize}

\end{comment}

%%%%%%%%%%%%%%%%%%%%%%%%%%%%%%%%%%%%%%%%%%%%%%%%%%%%%%%%%%%%%%%%%%%%%%%%%%%%%%%%
%%%%%%%%%%%%%%%%%%%%%%%%%%%%%%%%%%%%%%%%%%%%%%%%%%%%%%%%%%%%%%%%%%%%%%%%%%%%%%%%

\vspace{-1mm}
\section{Synthetic Datasets}\label{sec:datasets}
\vspace{-1mm}

In this section, we describe the synthetic datasets used in our library.
We start by discussing the benefits of synthetic datasets when evaluating feature
attribution methods, and then describe the feature distributions implemented for these datasets.

\vspace{-1mm}
\subsection{The case for synthetic data}
\vspace{-1mm}

As shown in Section \ref{sec:metrics}, for multiple metrics it is key to compute the conditional expectation 
$\E_{\bm{x}'\sim \D\left(\bm{x}_{S}\right)}[f(\bm{x}')]$ for a subset $S$, datapoint $\bm{x}$,
and trained model $f$. On real-world datasets, the conditional distribution
$\D\left(\bm{x}_{S}\right)$ can only be approximated, and the approximation
may be very poor when the conditional distribution
defines low-density regions of the feature space. Since all evaluation metrics require
computing $\Theta(D)$ or $\Theta(2^D)$ expectations for each datapoint $\bm{x}$,
is is likely that some evaluations will make use of a poor approximation.
However, for the synthetic datasets that we define, the conditional distributions are known,
allowing exact computation of the evaluation metrics.

Additionally, as we show in Section \ref{sec:experiments}, synthetic datasets allow one 
to explicitly control all attributes of the dataset, which allows for targeted experiments,
for example, investigating explainer performance as a function of feature correlation.
For explainers such as SHAP~\citep{shap} which assume feature independence, this type of
experiment may be very beneficial.
Finally, synthetic datasets can be used to simulate real datasets, which enables fair 
benchmarking of explainers with quantitative metrics.

%In addition, synthetic datasets offer direct handle to control the degree of correlation between features, and the generative model of producing targets from features, allowing us to benchmark explainers in a wide range of scenarios.

%Finally, the modular design of feature and target generators facilitates expandability in future work.

%No privacy concerns

\vspace{-1mm}
%\subsection{Mutivariate Gaussian and mixture of Gaussians features}
\subsection{Synthetic feature distributions}
\vspace{-1mm}

Now we describe the synthetic datasets in our library.
In general, the datasets are expressed as $ y = h(\bm{x}) $, with y as 
label and $\bm{x}$ as feature vector. 
The generation is split into two parts, generating features $\bm{x}$, 
and defining a function to generate labels $y$ from $\bm{x}$.
We implement multiple families of synthetic distributions in our library,
including multivariate Gaussian, mixture of Gaussians, and multinomial feature distributions.

%\subsubsection{Multivariate Gaussian}
To give a concrete example, we describe here how to generate and use
% We first describe feature generation, beginning with
multivariate Gaussian synthetic features.
The multivariate normal distribution of a  $D$-dimensional random vector \(\boldsymbol X = (X_1, ... , X_D)^T \) can be written as
    \(\boldsymbol X \sim \mathcal{N}(\boldsymbol \mu,\, \boldsymbol \Sigma),\)
where $\boldsymbol \mu$ is the $D$-dimensional mean vector, and $\boldsymbol \Sigma$ is the $D \times D$ covariance matrix. Without loss of generality, we can partition the $D$-dimensional vector $\boldsymbol{x}$ as 
    $\boldsymbol{X} = (\boldsymbol X_1,\boldsymbol X_2 )^T.$
To compute the distribution of $\boldsymbol X_1$ conditional on $\boldsymbol{X_2 = x_2^*}$ where $\boldsymbol{x_2^*}$ is a $K$-dimensional vector with $0 < K < D$, we can then partition $\boldsymbol \mu$ and $\boldsymbol \Sigma$ accordingly:
\begin{align*}
    \boldsymbol{\mu} = \begin{bmatrix} \boldsymbol \mu_1 \\ \boldsymbol \mu_2 \end{bmatrix}, 
    \hspace{4mm}
    \boldsymbol{\Sigma} = \begin{bmatrix} \boldsymbol \Sigma_{11} && \boldsymbol \Sigma_{12}\\ \boldsymbol \Sigma_{21} && \boldsymbol \Sigma_{22} \end{bmatrix}.
\end{align*}
Then the conditional distribution is a new multivariate normal \( (\boldsymbol X_1 | \boldsymbol X_2 = \boldsymbol x_2^*) \sim \mathcal{N}(\boldsymbol \mu^*,\, \boldsymbol \Sigma^*)\) where
\begin{equation} \label{eq:mu_and_sigma}
    \boldsymbol{\mu}^* = \boldsymbol \mu_1 + \boldsymbol \Sigma_{12} \boldsymbol \Sigma_{22}^{-1}(\boldsymbol x_2^* - \boldsymbol \mu_2),
    \hspace{4mm}
    \boldsymbol{\Sigma}^* = \boldsymbol \Sigma_{11} + \boldsymbol \Sigma_{12} \boldsymbol \Sigma_{22}^{-1}\boldsymbol \Sigma_{21}.
\end{equation}

For any $\boldsymbol x_2^* \in \mathbb{R}^K,$ one can
compute $\boldsymbol \mu^*$ and $\boldsymbol \Sigma^*$ and then generate samples from the conditional distribution.
Parameter $\boldsymbol \mu$ can take any value, and $\boldsymbol \Sigma$ must be symmetric and positive definite.
Similarly, we also give the derivation for additional distribution families in Appendix \ref{sec:mog}, including mixtures of multivariate Gaussians, and multinomial features.

\vspace{-1mm}
\subsection{Labels}
\vspace{-1mm}

After defining a distribution of features via one of the above distribution families,
we can then define a distribution over labels.
The distributions we implement are \texttt{linear}, \texttt{piecewise constant},
\texttt{nonlinear additive}, and \texttt{piecewise linear}.

Data labels are computed in two steps: \emph{(1)} raw labels are computed from features,
i.e. $y_\text{raw} = \sum_{n=1}^{D} \Psi_n(x_n)$
where $\Psi_n$ is a function that operates on feature $n$, and
\emph{(2)} final labels are normalized to have zero mean and unit variance.
The normalization ensures that a baseline ML model, which always predicts the mean of the dataset, has an MSE of 1.
This allows results derived from different types of datasets to be comparable at scale.

For \texttt{linear} datasets, $\Psi_n(x_n)$ are scalar weights, and we can rewrite the raw labels as $y_{raw} = \bm{w}^T \bm{x}$. In our experiments in Section \ref{sec:experiments}, we set $\bm{w}=[0,1,\dots,d-1]$.
\texttt{piecewise linear} datasets are similar to \texttt{linear}, but a different weight vector is used in different parts of the feature space. In our experiments in Section \ref{sec:experiments}, on the datasets with continuous features, we set $\bm{w}=[0,1,\dots,d-1]$ when the sum of the feature values is positive, and $\bm{w}=[d-1, d-2, \dots, 0]$ otherwise.
For \texttt{piecewise constant} datasets, $\Psi_n(x_n)$ are piecewise constant functions 
made up of different threshold values (similar to Aas et al.~\citep{shapr}).
For \texttt{nonlinear additive} datasets, $\Psi_n(x_n)$ are nonlinear functions including \textit{absolute}, \textit{cosine}, and \textit{exponent} function adapted from Chen et al.~\citep{l2x}.
Detailed specifications can be found in Appendix \ref{sec:dataset_details}.

%% file: sections/5-experiments.tex
\vspace{-1mm}
\section{Experiments}\label{sec:experiments}
\vspace{-1mm}

% In this section, 
% we run experiments with several popular feature
% attribution methods across synthetic datasets.
We show experiments on several popular feature
attribution methods across synthetic datasets.

\vspace{-1mm}
\subsection{Feature attribution methods}
\vspace{-1mm}
We compare eight different feature attribution methods:
SHAP~\citep{shap}, SHAPR~\citep{shapr}, brute-force Kernel SHAP (BF-SHAP)~\citep{shap},
LIME~\citep{lime}, MAPLE~\citep{maple}, L2X~\citep{l2x}, breakDown~\citep{breakdown},
and the baseline RANDOM, which outputs random weights drawn from a standard normal distribution.
We ran light hyperparameter tuning on all datasets.
See Appendix~\ref{app:descriptions} for details and descriptions 
for all methods. We report the mean and standard deviation 
from ten trials for all experiments.

\vspace{-1mm}
\subsection{Parameterized synthetic data experiments} \label{sec:parameterized}
\vspace{-1mm}
We first show experiments using multivariate Gaussian datasets described in Section~\ref{sec:datasets}. 
Without loss of generality, we can assume that the feature set is normalized (in other words,
$\boldsymbol \mu$ is set to 0, and the diagonal of $\boldsymbol \Sigma$ is set to 1).
In all sections except Section~\ref{subsec:wine}, we set the non-diagonal terms of $\boldsymbol \Sigma$ 
to $\boldsymbol \rho$, which allows for the convenient parameterization of a global level of feature 
dependence~\cite{shapr}. 

We run experiments that compare eight feature attribution methods on the five evaluation
metrics defined in Section~\ref{sec:prelim} across several datasets and ML models.
We conduct experiments by varying one or two of these dimensions at a time
while holding the other dimensions fixed (for example, we compare different datasets while keeping
the ML model fixed) and in Appendix~\ref{sec:additional}, we give the exhaustive set
of experiments.
Throughout this section, we will identify different types of failure modes, for example,
failures for some explainability techniques over specific metrics 
(Table~\ref{tab:results_metrics}) or failures for some techniques on datasets with
high levels of feature correlation 
(Figures~\ref{fig:vary_dataset_and_rho} and~\ref{fig:vary_ml_model}).

% todo: update hte discussion of experimental results

\vspace{-1mm}
\paragraph{Performance across metrics}
As shown in Table~\ref{tab:results_metrics}, the relative performance of explainers varies dramatically across metrics for a fixed 
multilayer perceptron
trained on a \texttt{nonlinear additive} dataset with $\rho=0.5$.  
Since $\rho=0.5$ implies that the features are fairly correlated, we find that SHAPR outperforms SHAP on GT-Shapley, which is consistent with the fact that SHAPR was designed to outperform SHAP in the presence of dependent features~\citep{shapr}. SHAPR achieved the top performance for three metrics, but MAPLE had the most consistent performance across all five metrics. One possible explanation for this is that MAPLE draws on ideas from three different areas of explainability: example-based, local, and global explanations~\citep{maple}, which helps it achieve steady performance across many metrics. Finally, while breakDown achieves the worst score for GT-Shapley, it achieves the best score for monotonicity. Note that breakDown works by greedily choosing the features with the greatest effect on the model output, \emph{with replacement}, making it particularly well-suited for the monotonicity metric, which checks whether replacing features sorted by importance with their background value with replacement monotonically decreases the change in model output.

\input{tables/results_metrics}

\vspace{-1mm}
\paragraph{Performance across dataset types and feature correlations}
Next, we explore how the type of dataset and feature correlation affects performance of explainers on a multilayer perceptron with the faithfulness metric. As shown in Figure \ref{fig:vary_dataset_and_rho}, a general trend is that explainers become less faithful as feature correlation increases. 
Explainers such as Kernel SHAP assume feature independence~\citep{shapr, molnar2019} and tend to perform well when features are indeed independent ($\rho = 0$). This is especially apparent with the \texttt{linear} dataset, where the performance of most methods cluster above 0.9 at $\rho=0$. However, LIME's performance drops as much as $\sim 90\%$ when features are almost perfectly correlated ($\rho = 0.99$). On the other hand, for both
the \texttt{nonlinear additive} and \texttt{piecewise constant} datasets, 
MAPLE's performance stayed relative stable across
values of $\rho$.
For experiments on the \texttt{piecewise linear} dataset, see Appendix \ref{sec:additional}.

\input{figures/vary_dataset_and_rho}

\vspace{-1mm}
\paragraph{Performance across ML models}
Next, we train three ML models: linear regression, decision tree, and multilayer perceptron, with a \texttt{piecewise constant} dataset and compare faithfulness. 
Figure \ref{fig:vary_ml_model} shows that
as in Figure~\ref{fig:vary_dataset_and_rho}, explainer performance drops as features become more correlated. 
Most explainers perform well for linear regression up to $\rho = 0.75$. The performance of SHAP, SHAPR, and LIME remain relatively consistent across ML models. In contrast, MAPLE performs significantly worse on the decision tree model.

\input{figures/vary_ml_model}

\subsection{Simulating real datasets} \label{subsec:wine}

In this section, we demonstrate the power and flexibility of synthetic datasets by simulating two popular datasets: the wine quality dataset \citep{wine,breakdown} and the forest fire dataset \citep{forestfire} with synthetic features so that they can be used to efficiently benchmark feature attribution methods.

\paragraph{Wine quality dataset}
The wine dataset has 11 continuous features ($\bm{x}_\text{real}$) and one integer quality rating ($y_\text{real}$) between 0 and 10.
In this section, it is formulated as a regression task, but it can also be formulated as a multi-class classification task.
The features are first normalized to have zero mean and unit variance, then an empirical covariance matrix is computed (Appendix Figure \ref{fig:wine_cov}), which is then used as the 
input covariance matrix to generate synthetic multivariate Gaussian features ($\bm{x}_\text{sim}$).
Simulated wine quality ($y_\text{sim}$) is labeled by a $k$-nearest neighbor model based on real datapoints ($\bm{x}_\text{real}, y_\text{real}$). 

We evaluate how close the simulated dataset is to the real one in two steps. First, we compute the Jensen-Shannon Divergence (JSD)~\citep{wong1985entropy} 
of the real and synthetic wine datasets. JSD measures the similarity between two distributions; it is bounded between 0 and 1, and lower JSD suggests higher similarity between two distributions. 
The JSD of marginal distributions between the real empirical features and the synthetic Gaussian features has a mean of $0.20$, and the JSD of real and synthetic targets is $0.23$, suggesting a good fit. 
Second, we train three types of ML models on both simulated and real wine datasets and compare the
MSE of explanations on a common held-out real test set. As shown in Appendix Table \ref{tab:real_sim_mse}, consistent low MSE across ML models and explainers suggest that the simulated dataset is a good proxy for the original wine dataset for evaluating explainers. 

Next, we compute evaluation metrics for seven different explainers on the synthetic wine dataset.
Note that computing these metrics accurately is not possible on the real wine dataset, as the conditional distribution is unknown. 
As shown in Table \ref{tab:wine_results}, SHAPR performs well on GT-Shapley, 
consistent with Table \ref{tab:results_metrics}. 
SHAP and SHAPR both outperform LIME and MAPLE on faithfulness.

\paragraph{Forest fire dataset}
The forest fire dataset has 12 continuous features and one real-valued label indicating the area of burned forest. Again, we normalize the features to have zero mean and unit variance, and then we compute the covariance matrix, which is used to generate the synthetic dataset (the same way as the wine quality dataset above).

For the forest fire dataset, the JSD of marginal distributions between the real empirical features and the synthetic Gaussian features has a mean of $0.17$, and the JSD of real and synthetic targets is $0.15$, suggesting a good fit. 
We compute evaluation metrics for six different explainers on the synthetic forest fire dataset. See Table \ref{tab:forestfires_results}. 
SHAP achieved top performance on three of the five metrics.

\input{tables/wine_results}
\input{tables/forestfires}

\subsection{Recommended usage}
In Section~\ref{sec:experiments}, we gave a sample of the types of experiments
that can be performed with our library~(recall that comprehensive experiments are in
Appendix~\ref{sec:additional}).
For researchers looking to develop new explainability techniques, we recommend
benchmarking new algorithms across all metrics using our synthetic datasets with different values of $\rho$. 
These datasets give a good initial picture of the efficacy of new techniques.
For researchers with a dataset and application in mind, we recommend converting the dataset
into a synthetic dataset using the technique described in Section~\ref{subsec:wine}.
Note that converting to a synthetic dataset also gives the ability to evaluate explainability
techniques on perturbations of the original covariance matrix, to simulate robustness
to distribution shift.
Finally, researchers can decide on the evaluation metric that is most suitable to the
application at hand.
See Section~\ref{subsec:guide} for a guide to choosing the best metric based on the 
application.

%% file: tables/results_metrics.tex
\begin{table}[h]
    \vspace{-3mm}
    \caption{Explainer performance across metrics. All performance numbers are from 
    explaining a multilayer perceptron trained on the Gaussian nonlinear additive dataset with $\rho = 0.5$. 
    } 
    \label{tab:results_metrics}
    \vspace{2mm}
\resizebox{\textwidth}{!}{
\begin{tabular}{cccccccc}
\hline
  & RANDOM            & SHAP           & SHAPR             & LIME              & MAPLE              & L2X & BREAKDOWN              \\ \hline
faithfulness($\uparrow$)     & $0.002_{\pm0.034}$    & $0.651_{\pm0.051}$    & $\textbf{0.799}_{\pm0.036}$    & $0.524_{\pm0.06}$    & $0.478_{\pm0.061}$    & $0.000_{\pm0.075}$    & $0.110_{\pm0.049}$ \\

monotonicity($\uparrow$)    & $0.525_{\pm0.017}$    & $0.537_{\pm0.014}$    & $0.550_{\pm0.025}$    & $0.517_{\pm0.022}$    & $0.543_{\pm0.026}$    & $0.535_{\pm0.022}$    & $\textbf{0.562}_{\pm0.021}$       \\ 

ROAR($\uparrow$)  & $0.380_{\pm0.051}$    & $0.455_{\pm0.054}$    & $\textbf{0.465}_{\pm0.054}$    & $0.432_{\pm0.051}$    & $0.432_{\pm0.059}$    & $0.365_{\pm0.053}$    & $0.329_{\pm0.057}$    \\ 

GT-Shapley($\uparrow$)   & $0.004_{\pm0.049}$    & $0.810_{\pm0.023}$    & $\textbf{0.930}_{\pm0.012}$    & $0.711_{\pm0.032}$    & $0.530_{\pm0.128}$    & $-0.014_{\pm0.068}$    & $-0.127_{\pm0.066}$   \\

infidelity($\downarrow$) & $0.114_{\pm0.058}$    & $0.050_{\pm0.023}$    & $0.036_{\pm0.013}$    & $0.053_{\pm0.016}$    & $\textbf{0.019}_{\pm0.011}$    & $0.025_{\pm0.010}$    & $0.126_{\pm0.057}$ \\
\hline
\end{tabular}}
\vspace{-1mm}
\end{table}

%% file: figures/vary_dataset_and_rho.tex
\begin{figure}[t]
    \begin{center}
        \includegraphics[width=\linewidth]{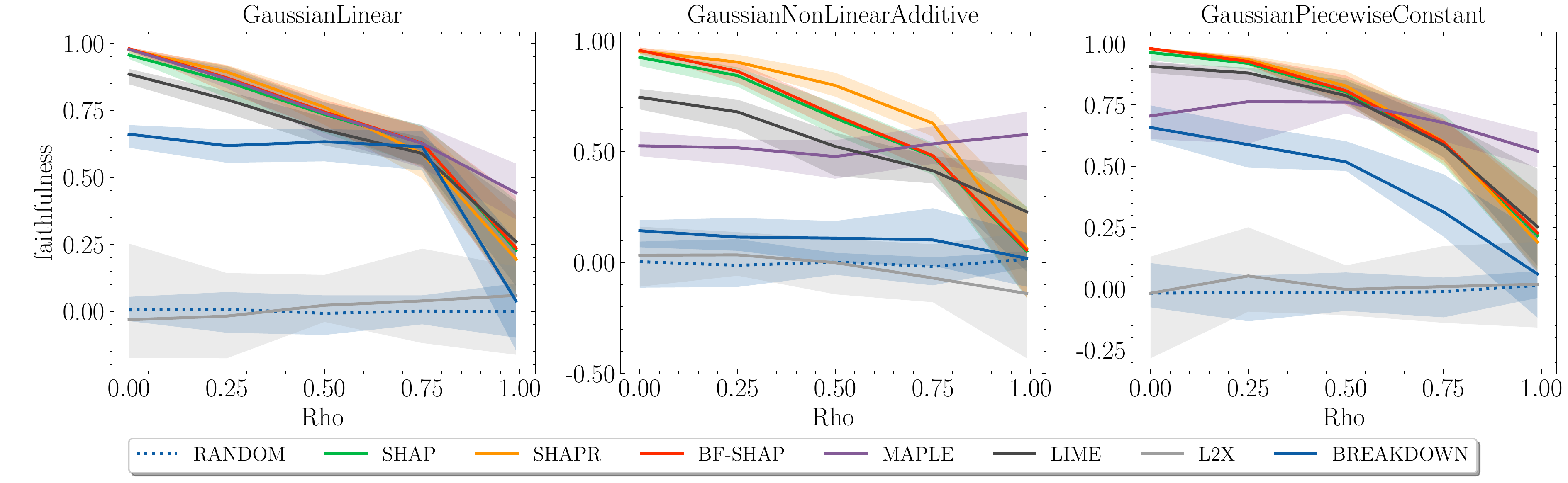}
    \end{center}
    \vspace{-1mm}
    \caption{Results for faithfulness on a multilayer perceptron trained on three 
    different datasets.}
%    types of Gaussian datasets.}
    \label{fig:vary_dataset_and_rho}
    \vspace{-3mm}
\end{figure}

%% file: figures/vary_ml_model.tex
\begin{figure}[t]
    \begin{center}
        \includegraphics[width=\linewidth]{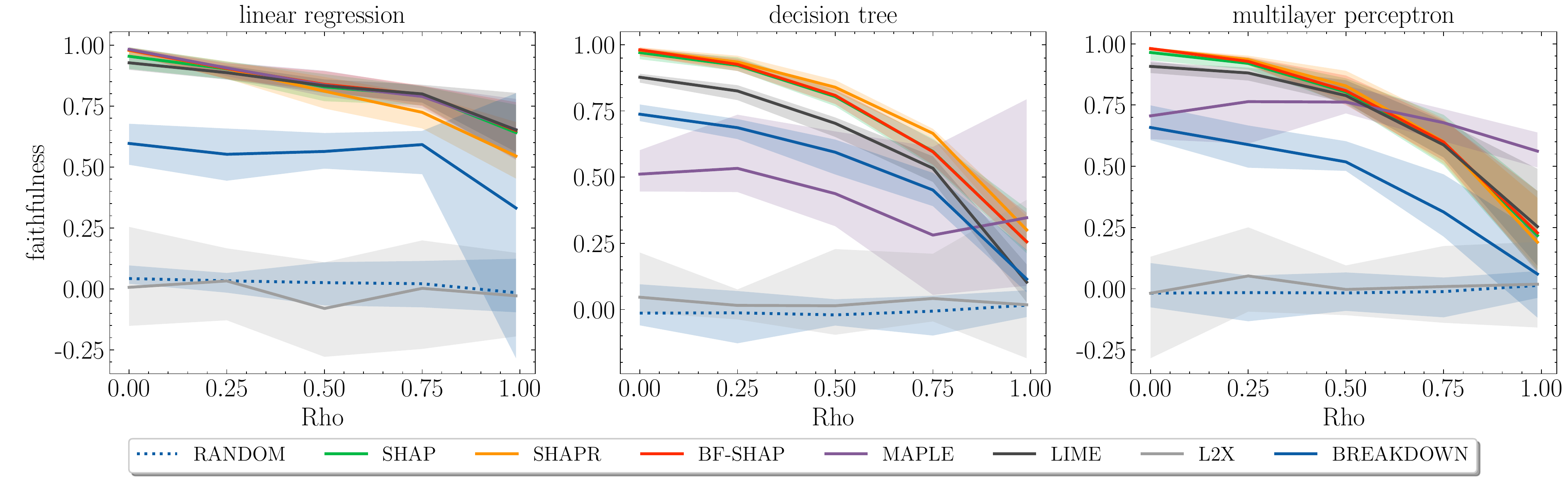}
    \end{center}
    \vspace{-1mm}
    \caption{Results for faithfulness for three types of ML models---linear regression, 
    decision tree, and multilayer perceptron---trained on a Gaussian piecewise constant dataset.}
    \label{fig:vary_ml_model}
    \vspace{-4mm}
\end{figure}

%% file: tables/wine_results.tex
\begin{table}[h]
    \caption{Explainer performance on the simulated wine dataset across metrics. 
    All performance numbers are from explainers for a decision tree.}
    \label{tab:wine_results}
    \vspace{1mm}
    \centering
    \resizebox{\textwidth}{!}{
    \begin{tabular}{c c c c c c c c}
        \hline
         & RANDOM & SHAP & SHAPR & LIME & MAPLE & L2X & BREAKDOWN \\
        \hline
        
        faithfulness ($\uparrow$)
        & $-0.007_{\pm0.005}$    & $\textbf{0.534}_{\pm0.045}$ &  $0.528_{\pm 0.032}$  & $0.368_{\pm0.031}$    & $0.034_{\pm0.033}$    & $-0.030_{\pm0.018}$    & $-0.042_{\pm0.011}$
        \\
        monotonicity ($\uparrow$)
        & $0.529_{\pm0.008}$    & $0.549_{\pm0.009}$ &  $\textbf{0.551}_{\pm 0.009}$  & $0.547_{\pm0.007}$    & $0.520_{\pm0.014}$    & $0.522_{\pm0.005}$    & $0.493_{\pm0.014}$  
        \\
        ROAR ($\uparrow$)
        & $0.698_{\pm0.031}$    & $0.780_{\pm0.016}$ &  $0.549_{\pm 0.031}$  & $0.738_{\pm0.026}$    & $\textbf{0.818}_{\pm0.022}$    & $0.664_{\pm0.02}$    & $0.625_{\pm0.002}$ 
        \\
        GT-Shapley ($\uparrow$)
        & $0.004_{\pm0.013}$    & $0.825_{\pm0.006}$ &  $\textbf{0.945}_{\pm0.002}$  & $0.745_{\pm0.015}$    & $0.685_{\pm0.008}$    & $-0.108_{\pm0.029}$    & $-0.064_{\pm0.02}$  
        \\
        infidelity ($\downarrow$)
        & $0.353_{\pm0.174}$    & $0.234_{\pm0.124}$ &  $\textbf{0.212}_{\pm0.146}$  & $0.234_{\pm0.126}$    & $0.234_{\pm0.132}$    & $0.285_{\pm0.115}$    & $0.365_{\pm0.133}$     
        \\
        \hline
    \end{tabular}}
    \vspace{1mm}
\end{table}

%% file: tables/forestfires.tex
\begin{table}[h]
    \caption{Explainer performance on the simulated forest fires dataset across metrics. 
    All performance numbers are from explainers for a decision tree.}
    \label{tab:forestfires_results}
    \vspace{1mm}
    \centering
    \resizebox{\textwidth}{!}{
    \begin{tabular}{c c c c c c c c}
        \hline
         & RANDOM & SHAP & LIME & MAPLE & L2X & BREAKDOWN \\
        \hline
        
        faithfulness ($\uparrow$)
        & $0.022_{\pm0.034}$    & $\textbf{0.571}_{\pm0.023}$    & $0.449_{\pm0.007}$    & $0.080_{\pm0.056}$    & $0.001_{\pm0.008}$    & $0.158_{\pm0.032}$     
        \\
        monotonicity ($\uparrow$)
        & $0.537_{\pm0.02}$    & $0.591_{\pm0.007}$    & $\textbf{0.598}_{\pm0.002}$    & $0.561_{\pm0.002}$    & $0.527_{\pm0.01}$    & $0.575_{\pm0.012}$     
        \\
        ROAR ($\uparrow$)
        & $0.575_{\pm0.002}$    & $0.615_{\pm0.011}$    & $0.616_{\pm0.008}$    & $\textbf{0.696}_{\pm0.024}$    & $0.534_{\pm0.018}$    & $0.604_{\pm0.019}$
        \\
        GT-Shapley ($\uparrow$)
        & $0.012_{\pm0.06}$    & $\textbf{0.870}_{\pm0.005}$    & $0.779_{\pm0.027}$    & $0.804_{\pm0.011}$    & $0.031_{\pm0.12}$    & $0.105_{\pm0.013}$  
        \\
        infidelity ($\downarrow$)
        & $0.207_{\pm0.125}$    & $\textbf{0.075}_{\pm0.074}$    & $0.077_{\pm0.075}$    & $0.077_{\pm0.079}$    & $0.091_{\pm0.07}$    & $0.117_{\pm0.076}$     
        \\
        \hline
    \end{tabular}}
    \vspace{1mm}
\end{table}

%% file: sections/6-impact.tex
\vspace{-2mm}
\section{Societal Impact}\label{sec:impact}
\vspace{-1mm}

Machine learning models are more prevalent now than ever before. 
With the widespread deployment of models in applications that impact human
lives, explainability is becoming increasingly important for the purposes of
debugging, legal obligations, and mitigating 
bias~\citep{shap, zhang2020survey, arya2019one, deyoung2019eraser}.
Given the importance of high-quality explanations, it is essential that
explainability methods are reliable across all types of datasets.
Our work seeks to speed up the development of explainability methods,
with a focus on catching edge cases and failure modes, to ensure
that new explainability methods are robust before they are used in the real 
world. Of particular importance are improving the reliability of
explainability methods intended to recognize biased predictions,
for example, ensuring that the features used to predict criminal
recidivism are not based on race or gender~\citep{larson2016we}.
Frameworks for evaluating and comparing explainability methods
are an important part of creating inclusive and unbiased technology.
As pointed out in prior work~\citep{denton2019detecting},
while methods for explainability or debiasing are important,
they must be part of a larger, socially contextualized project
to examine the ethical considerations of the machine learning application.

%% file: sections/10-conclusion.tex
\vspace{-2mm}
\section{Conclusions and Limitations}\label{sec:conclusion}
\vspace{-1mm}
% talking about limitations is one of the requirements for neurips 2021
In this work, we released a set of synthetic datasets along with a library for benchmarking
feature attribution algorithms. The use of synthetic datasets with known ground-truth 
distributions makes it possible to exactly compute the conditional distribution over any set
of features, enabling accurate computations of several explainability evaluation metrics, including
ground-truth Shapley values, ROAR, faithfulness, and monotonicity.
Our synthetic datasets offer a variety of parameters which can be configured to simulate
real-world data and have the potential to identify failure modes of explainability techniques,
for example, techniques whose performance is negatively correlated with dataset feature
correlation.
We showcase the power of our library by benchmarking several popular explainers
with respect to five evaluation metrics across a variety of settings.
%Following initial feedback on our work, we reduced the number of metrics to include only the most popular and useful metrics (described in Section~\ref{app:statement}).

Despite the fact that the synthetic datasets aim to cover a broad range of feature 
distributions, correlations, scales, and target generation functions, there is almost certainly a gap 
between synthetic and real-world datasets. However, as discussed before, it is often the case that we do 
not know the ground truth generative model of real datasets, thus making it impossible to compute 
many objective metrics. Hence, there is a trade-off between data realism and ground truth availability. 

Note that our library is \textbf{not} meant to be a replacement for human interpretability studies. 
Since the goals of explainability methods are inherently human-centric, the only foolproof method
of evaluating explanation methods are to use human trials. Rather, our library is meant to substantially
speed up the process of development, refinement, and identifying failures, 
before reaching human trials.

Overall, we recommend developing new explainability methods in this library, and then
conducting human trials on real data.
Our library is designed to substantially accelerate the process of moving 
new explainability algorithms from development to deployment.
With the release of API documentation, walkthroughs, and a contribution guide, we hope that the scope
of our library can increase over time.

\begin{comment}

The versatility and efficiency of our library will help researchers bring their explainability methods 
from development to deployment.

All of our code, API docs, and walkthroughs are available at 
\url{https://github.com/abacusai/xai-bench}.

\end{comment}

%% file: sections/99-appendix.tex
\section{Dataset Documentation and Intended Use} \label{app:documentation}

Our code is available at \url{https://github.com/abacusai/xai-bench}.

\subsection{Author responsibility}
We bear all responsibility in case of violation of rights, etc.
The license of our repository is the
\textbf{Apache License 2.0}. For more information,
see \url{https://github.com/abacusai/xai-bench/blob/main/LICENSE}.

\subsection{Maintenance plan and contributing policy.}
We plan to actively maintain the repository, and we welcome contributions from the explainability
community and machine learning community at large.
For more information, see
\url{https://github.com/abacusai/xai-bench}.
As our benchmarks are synthetic, we will host the code to generate the datasets on GitHub.

\subsection{Code of conduct}
Our Code of Conduct is adapted from the Contributor Covenant, version 2.0, 
available at \\
\url{https://www.contributor-covenant.org/version/2/0/code_of_conduct.html}. \\
The policy is copied below.

\begin{quote}
    ``We as members, contributors, and leaders pledge to make participation in our community a harassment-free experience for everyone, regardless of age, body size, visible or invisible disability, ethnicity, sex characteristics, gender identity and expression, level of experience, education, socio-economic status, nationality, personal appearance, race, caste, color, religion, or sexual identity and orientation.''
\end{quote}

\section{Reproducibility Checklist}
To ensure reproducibility, we use the Machine Learning Reproducibility Checklist v2.0, Apr.\ 7, 
2020~\citep{pineau2020improving}. An earlier verision of this checklist (v1.2) was used 
for NeurIPS 2019~\citep{pineau2020improving}.
% https://www.cs.mcgill.ca/~jpineau/ReproducibilityChecklist.pdf

\begin{itemize}
    \item For all \textbf{models} and \textbf{algorithms} presented,
    \begin{itemize}
        \item \textbf{A clear description of the mathematical setting, algorithm, and/or model.}
        We clearly describe all of the settings and algorithms in Section~\ref{sec:prelim}
        and Appendix Section~\ref{app:descriptions}.
        \item \textbf{A clear explanation of any assumptions.}
        Some of the explainability techniques implemented in our repository make assumptions about
        the dataset (e.g., that all features are independent). We give this information in
        Appendix~\ref{app:descriptions}.
        \item \textbf{An analysis of the complexity (time, space, sample size) of any algorithm.}
        We reported the complexity analysis in Section~\ref{sec:prelim}
        and Appendix Section~\ref{app:descriptions}.
    \end{itemize}
    \item For any \textbf{theoretical claim},
    \begin{itemize}
        \item \textbf{A clear statement of the claim.} We do not make theoretical claims.
        \item \textbf{A complete proof of the claim.} We do not make theoretical claims.
    \end{itemize}
    \item For all \textbf{datasets} used, check if you include:
    \begin{itemize}
        \item \textbf{The relevant statistics, such as number of examples.}
        We used a real dataset in Section~\ref{subsec:wine}. We give the statistics for this
        dataset in the same section.
        \item \textbf{The details of train / validation / test splits}
        We give this information in our repository.
        \item \textbf{An explanation of any data that were excluded, and all pre-processing step.}
        We did not exclude any data or perform any preprocessing.
        \item \textbf{A link to a downloadable version of the dataset or simulation environment.}
        Our repository contains all of the instructions to download and run experiments on
        the datasets in our work. See \url{https://github.com/abacusai/xai-bench}.
        \item \textbf{For new data collected, a complete description of the data collection process, such
        as instructions to annotators and methods for quality control.}
        We release new synthetic datasets, so there was no collection process. The code to generate
        the synthetic datasets is hosted on GitHub.
    \end{itemize}
    \item For all shared \textbf{code} related to this work, check if you include:
    \begin{itemize}
        \item \textbf{Specification of dependencies.}
        We give installation instructions in the README of our repository.
        \item \textbf{Training code.}
        The training code is available in our repository.
        \item \textbf{Evaluation code.}
        The evaluation code is available in our repository.
        \item \textbf{(Pre-)trained model(s).}
        We do not release any pre-trained models. The code to run all experiments in our
        work can be found in the GitHub repository.
        \item \textbf{README file includes table of results accompanied by precise command to run to
        produce those results.}
        We include a README with detailed instructions to reproduce our experiments.
    \end{itemize}
    \item For all reported \textbf{experimental results}, check if you include:
    \begin{itemize}
        \item \textbf{The range of hyper-parameters considered, method to select the best 
        hyper-parameter configuration, and specification of all hyper-parameters used to 
        generate results.}
        We use default configuration for explainers except SHAPR, which we discuss in 
        Appendix~\ref{app:explainers}. Our repository allows setting the hyperparameters to other
        values set by the user.
        \item \textbf{The exact number of training and evaluation runs.}
        We reported that we ran ten trials for each experiment.
        \item \textbf{A clear definition of the specific measure or statistics used to report results.}
        We define our metrics in Section~\ref{sec:metrics}. 
        \item \textbf{A description of results with central tendency (e.g. mean) \& variation 
        (e.g. error bars).}
        We report mean and standard deviation for all experiments.
        \item \textbf{The average runtime for each result, or estimated energy cost.}
        We report the runtimes in Section~\ref{sec:additional}.
        \item \textbf{A description of the computing infrastructure used.}
        We use CPUs for all experiments. We give details of our experiments in 
        Appendix Section~\ref{sec:additional}.
    \end{itemize}
\end{itemize}
\section{Multivariate Gaussian distribution} \label{sec:gaussian}

The probability density function of a non-degenerative multi-variate normal distribution is
\begin{equation} \label{eq:normal}
  f_x(x_1,...,x_D) = \frac{\exp(-\frac{1}{2} {(\boldsymbol x-\boldsymbol\mu)}^T \boldsymbol\Sigma^{-1} (\boldsymbol x - \boldsymbol \mu))}{\sqrt{(2 \pi)^D |\boldsymbol \Sigma|}},
\end{equation}
with parameters $\bm \mu \in \mathbb{R}^D$ and 
$\boldsymbol \Sigma \in \mathbb{R}^{D \times D}$.

\section{Additional Synthetic Feature Distributions} \label{sec:mog}

\paragraph{Mixture of multivariate Gaussians features}
We first describe mixture of multivariate Gaussians features.
Suppose now that $\boldsymbol X = (X_1, ... , X_D)^T$ is a 
$D$-dimensional random vector distributed as a 
mixture of $k$ Gaussians. We write this as
$\boldsymbol X \sim
\sum_{j=1}^k \pi_j \mathcal{N}(\boldsymbol \mu_j, \boldsymbol \Sigma_j)$,
where each $\boldsymbol \mu_j$ is a $D$-dimensional mean vector for
the $j^\text{th}$ mixture component,
and $\boldsymbol \Sigma_j$ is the $D \times D$
covariance matrix for the $j^\text{th}$ mixture component.

Suppose, as before we use the partition defined by
$\boldsymbol{X} = \begin{bmatrix} \boldsymbol X_1 \\ \boldsymbol X_2 \end{bmatrix}$
and partition the parameters of each mixture component accordingly as
\begin{align*}
    \boldsymbol{\mu}_j =
    \begin{bmatrix}
        \boldsymbol \mu_{j, 1} \\ \boldsymbol \mu_{j, 2}
    \end{bmatrix}, 
    \boldsymbol{\Sigma}_j =
    \begin{bmatrix}
        \boldsymbol \Sigma_{j, 11} && \boldsymbol \Sigma_{j, 12}\\
        \boldsymbol \Sigma_{j, 21} && \boldsymbol \Sigma_{j, 22}
    \end{bmatrix}
\end{align*}
for $j=1,\ldots,k$.
Then, given $X_2 = \boldsymbol x_2^*$, the conditional distribution is 
also a mixture of Gaussians, written
$ (\boldsymbol X_1 | \boldsymbol X_2 = \boldsymbol x_2^*) \sim
\sum_{j=1}^k \pi_j^* \mathcal{N}(\boldsymbol \mu_j^*, \boldsymbol \Sigma_j^*)$,
where the parameters of each mixture component can be written

\begin{equation}
    \label{eq:mog_mu}
    \boldsymbol{\mu_j}^* =
    \boldsymbol \mu_{j, 1} + \boldsymbol \Sigma_{j, 12}
    \boldsymbol \Sigma_{j, 22}^{-1}(\boldsymbol x_2^* - \boldsymbol \mu_{j, 2})
\end{equation}

\begin{equation}
    \label{eq:mog_sigma}
    \boldsymbol{\Sigma_j}^* = 
    \boldsymbol \Sigma_{j, 11} + \boldsymbol \Sigma_{j, 12}
    \boldsymbol \Sigma_{j, 22}^{-1}\boldsymbol \Sigma_{j, 21}
\end{equation}

\begin{equation}
    \label{eq:mog_pi}
    \pi_j^* = 
    \frac{\pi_j f_{j, 2}(\boldsymbol x_2^*)}{
        \sum_{\ell=1}^k \pi_\ell  f_{\ell, 2}(\boldsymbol x_2^*)}
\end{equation}
and where $f_{j, 2}$ denotes the probability density function of
the multivariate normal distribution
$\mathcal{N}(\mu_{j, 2}, \Sigma_{j, 22})$.

\paragraph{Multinomial features}
We follow a similar derivation for the conditional distribution of a multinomial distribution.
Suppose now that $\boldsymbol X = (X_1, ... , X_D)^T$ is a 
$D$-dimensional random vector following a multinomial distribution,
where $X_i \in \{0, \ldots, m\}$, and $\sum_{i=1}^D X_i = m$.
We write this as
$\boldsymbol X \sim
\text{Multinomial}(m, p_1, \ldots, p_D)$,
where the parameter $m>0$ denotes the number of trials,
and the parameters $p_1,\ldots, p_D$ denote the $D$ event probabilities.

Suppose, as before we use the partition defined by
$\boldsymbol{X} = \begin{bmatrix} \boldsymbol X_1 \\ \boldsymbol X_2 \end{bmatrix}$.
Then, given $X_2 = \boldsymbol x_2^* \in \{0, \ldots, m\}^k$, the conditional distribution is 
also distributed as a multinomial, written
$ (\boldsymbol X_1 | \boldsymbol X_2 = \boldsymbol x_2^*) \sim
\text{Multinomial}(m^*, p_1^*, \ldots, p_{D-k}^*)$,
where the parameters of of this multinomial can be written
$m^* = m - \sum_{j=1}^k x_{2, j}^*$, and $p_i^* = p_i / \left(1 - \sum_{j-1}^k p_j\right)$.

\section{Descriptions of Explainability Metrics and Explainers}\label{app:descriptions}

\subsection{Metrics}\label{sec:appendix_metrics}

In this section, we give the formal definitions for the rest of the evaluation
metrics from Section~\ref{sec:evaluationmetrics}.
We start by giving the definition of the ROAR-based metrics.

Recall that the major difference between ROAR-based metrics and other metrics is that 
in order to evaluate the marginal improvement of sets of features, ROAR-based metrics
retrain the model using a new dataset with the features removed.
For example, rather than computing $\left|\E_{\bm{x}'\sim \D\left(\bm{x}_{F \setminus i}\right)}[f(\bm{x}')] -f(\bm{x})\right|$, we would compute
$\left|f^*(\E_{\bm{x}'\sim \D\left(\bm{x}_{F \setminus i}\right)}[\bm{x}']) -f(\bm{x})\right|$, where $f^*$ denotes a model that has been trained on a modification of $\Dtr$ where each datapoint has its $i$ features with highest weight removed.
Given a datapoint $\bm{x}$ and a set of features $S\subseteq F$, 
we start by defining $\bm{\bar x}_{S}$,
the expected value of a datapoint conditioned on the features $S$ from $\bm{x}$:

\begin{align}
    \bm{\bar x}_{S} &= 
    \begin{cases}
    &x_i \text{ for indices }i\in S\\
    &\E \left[x_i'\mid \bm{x'}\sim\D\text{ s.t. }x_j'=x_j\text{ for }j\in S\right]
    \text{ for indices }i\notin S
    \end{cases}
\end{align}

Recall from Section~\ref{sec:evaluationmetrics} that $S^+(\bm{w}, i)$ denotes the set of $i$ 
most important weights, and $S^+(\bm{w}, 0) = \emptyset$. 
%$S^+(\bm{w}, i)$ denotes the set of $i$ most important weights,
%
Let $\Dtr^{S(k)+}$ denote a new training set by replacing each $\bm{x}\sim \Dtr$ with
$\bm{\bar x}_{F\setminus S^+(\bm{w}(\bm{x}), k)}$, where $\bm{w}(\bm{x})$ denotes the weight
vector for $\bm{x}$. That is, $\Dtr^{k+}$ is the training set modified by removing
the $k$ most important features for each datapoint.
Let $f^{S(k)+}$ denote the model $f$ retrained on
$\Dtr^{S(k)+}$ instead of $\Dtr$.
Then \textbf{ROAR} is defined as follows:
\begin{align}
    \bar\delta_{i}^{+} \quad \quad &= \quad
    f^{S(k)+}(\bm{\bar x}_{S_{i+1}^{+}(\bm{w})})
    -f^{S(k)+}(\bm{\bar x}_{S_{i}^{+}(\bm{w})}),\\
    \text{ROAR} &= \quad
    \frac{1}{D-1}\sum_{i=0}^{D-2}                         \ind_{|\delta_{i}^{+}|\leq
    |\bar\delta_{i+1}^{+}|}
\end{align}

%%%%%%%%%%%%%%%%%%%%%%%%%%%%%%%%%%%%%%%%%%%%%%%%%%%%%%%%%%%%%%%%%%%%%%%%%%%%%%%%%%%%%
Now we give the formal definition for Shapley values.
Given a datapoint $\bm{x}$, the Shapley value $v_i$ is defined as follows.

\begin{equation}\label{eq:shapley}
    v_i = \sum_{S \subseteq F \setminus \{i\}}
        \frac{|S|!(|F|-|S|-1)!}{|F|!}
        (
        \E_{\bm{x}'\sim \D\left(\bm{x}_{S \cup \{i\}}\right)}[f(\bm{x}')]
%        \E[f_{S \cup \{i\}}(\bm{x}_{S \cup \{i\}})]
        -
        \E_{\bm{x}'\sim \D\left(\bm{x}_{S}\right)}[f(\bm{x}')]
%        \E[f_{S}(\bm{x}_{S})]
        ),
\end{equation}

where $\D\left(\bm{x}_{S}\right)$ is defined as in Equation~\ref{eq:conditional}.
Then for a datapoint $\bm{x}$, ground truth Shapley correlation 
is defined as the correlation between the
weight vector $\bm{w}$ and the set of Shapley values for $\bm{x}$.
%and shapley-mse is defined as the mean squared error between the weight vector $\bm{w}$ and the set of Shapley values for $\bm{x}$.
Formally,

\begin{equation}
    \text{GT-Shapley} = \texttt{Pearson}
        \left([v_i]_{1\leq i\leq D}, 
        [w_i]_{1\leq i\leq D}\right).
        %\\\text{shapley-mse} &= \sum_{i=1}^D \left(v_i - w_i\right)^2.
\end{equation}

The main drawback of this metric is its time complexity, which is $\Theta(2^D)$ for a $D$-dimensional dataset. Computation quickly becomes infeasible as $D$ scales up.

\subsection{Local Feature Attribute Explainers}\label{app:explainers}

In this section, we give descriptions and implementation details of all of the 
explainability methods and metrics implemented in our library.

\subsubsection{SHAP}

Lundberg et al.~\citep{shap} proposed a few methods such as BF-SHAP to estimate Shapley values defined by Equation~\ref{eq:shapley}. Due to the unavailability of the generative model of conditional distribution for real datasets, one can not accurately compute $\E[f_{S}(\bm{x}_{S})]$. BF-SHAP makes two assumptions: (1) model linearity, which makes $\E[f_{S}(\bm{x}_{S})] = f_{S}(\E[\bm{x}_{S}])$, (2) feature independence assumption: $\E[\bm{x}_{S}]$ with \textbf{marginal} expectation instead of \textbf{conditional} expectation. In this work, we refer the official implementation of SHAP as SHAP, and re-implemented brute-force kernel SHAP as BF-SHAP.

\subsubsection{SHAPR}

Aas et al.~\citep{shapr} proposes several techniques to relax both assumptions and improve BF-SHAP such as ``Gaussian'', ``copula'', and ``empirical''. Because the ``empirical'' method with a fixed $\sigma$ performs well across tasks in the original paper, we re-implemented the original R package in python with a tuned from \{0.1, 0.2, 0.4, 0.8\} and fixed $\sigma = 0.4$  and refer it as SHAPR. 

\subsubsection{LIME}

Local Interpretable Model-agnostic Explanations (LIME)~\citep{lime} interprets individual predictions based on locally approximating the model
around a given prediction. We use LIME from the official SHAP repository.

\subsubsection{MAPLE}

MAPLE~\citep{maple} is another technique that combines local neighborhood selection with local feature selection. We use official implementation from the official SHAP repository.

\subsubsection{L2X}
L2X~\citep{l2x} used a mutual information-based approach to explainability.
The L2X explainer has a hyperparameter $k$ which needs to be defined by the user to decide the top $k$ most important features to pick. For each $D$-dimensional data point, L2X outputs a $D$-dimensional binary vector $I_k$ with $1$ indicating important features and $0$ indicating unimportant features. Because $k$ is often unknown a priori, we modified L2X as follows:
\begin{equation}
    \bm{w} = \frac{2}{k(k+1)}\sum_{k=1}^{D}{I_k},
\end{equation}

where $\frac{2}{k(k+1)}$ is a scaling factor to ensure the elements in $\bm{w}$ sum up to $1$. The original L2X model uses 1 million training samples to achieve good performance, due to the computation limitation of metrics calculation, we limit the training set size of synthetic experiment to 1000, and experiments show that L2X often fails to achieve good performance.

\subsubsection{BREAKDOWN}
BREAKDOWN ~\cite{breakdown} is another technique  to decompose model predictions into parts that can be attributed to particular variables. We use the official python implementation from \url{https://github.com/MI2DataLab/pyBreakDown}.

\subsubsection{RANDOM}

RANDOM explainer is implemented to serve as a baseline model. The explainer generates random weights from standard normal distribution.

\section{Dataset details}\label{sec:dataset_details}

For 5-dimensional datasets, linear $w = [4,3,2,1,0]$, 

piecewise constant:
\begin{align}
    \Psi_{1}(x_1) &= 
    \begin{cases}
    &1, \;x_1>=0 \\
    &-1, \;x_1<0
    \end{cases}\\
    \Psi_{2}(x_2) &= 
    \begin{cases}
    &-2, \;x_2 < -0.5 \\
    &-1, \;-0.5 \leq x_2 < 0 \\
    &1, \;-0 \leq x_2 < 0.5 \\
    &2, \;x_2 \geq 0.5 \\
    \end{cases}\\
    \Psi_{3}(x_3) &= floor(2 cos(\pi x_3))\\
    \Psi_{i}(x_i) &= 0, \;i=4,5
\end{align}
where $floor()$ is a rounding function that rounds a real number to the nearest integer with the lowest absolute value.

Nonlinear additive:
\begin{align}
    \Psi_{1}(x_1) &= sin(x_1)\\
    \Psi_{2}(x_2) &= |x_2|\\
    \Psi_{3}(x_3) &= x_3^2\\
    \Psi_{4}(x_4) &= e^{x_4}\\
    \Psi_{5}(x_5) &= 0
\end{align}
where $floor()$ is a rounding function that rounds a real number to the nearest integer with lowest absolute value.

\input{figures/synthetic_dist}

\input{figures/wine_cov}

\section{Additional results} \label{sec:additional}

In this section, we present additional results and experimental details.

\input{tables/runtimes}

Table \ref{tab:runtimes} shows the time explainers take to generate explanations for 100 test datapoints. All of our experiments were run on CPUs. We report mean and standard deviation across three runs for all experiments except for Table~\ref{tab:runtimes}. 
All synthetic experiments have a training size of 1000, and test size of 100.

The wine dataset contains 4898 datapoints.
In Table~\ref{tab:real_sim_mse}, we give the mean squared error between explanations for
predictions of models trained on the real vs.\ simulated wine dataset described in
Section~\ref{sec:experiments}. 

\input{tables/real_sim_mse}

We conclude by presenting the comprehensive results for five different evaluation metrics, eight different feature attribution algorithms, nine different datasets, and five different values of $\rho$.

\input{figures/all_plots}

%% file: figures/synthetic_dist.tex
\begin{figure}[h]
    \begin{center}
    \begin{tabular}{ccc}
        \includegraphics[width=0.32\linewidth]{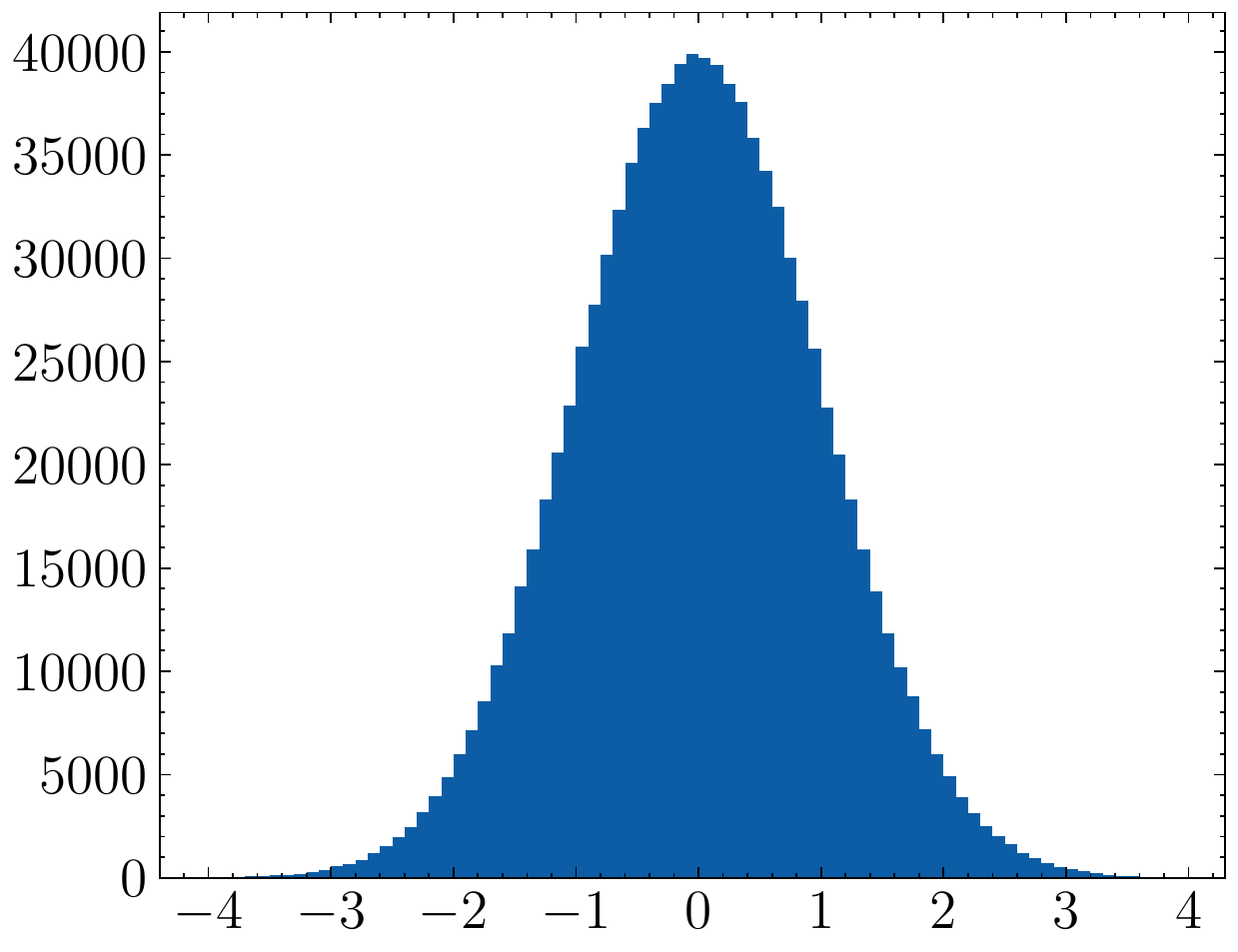}
        & \includegraphics[width=0.32\linewidth]{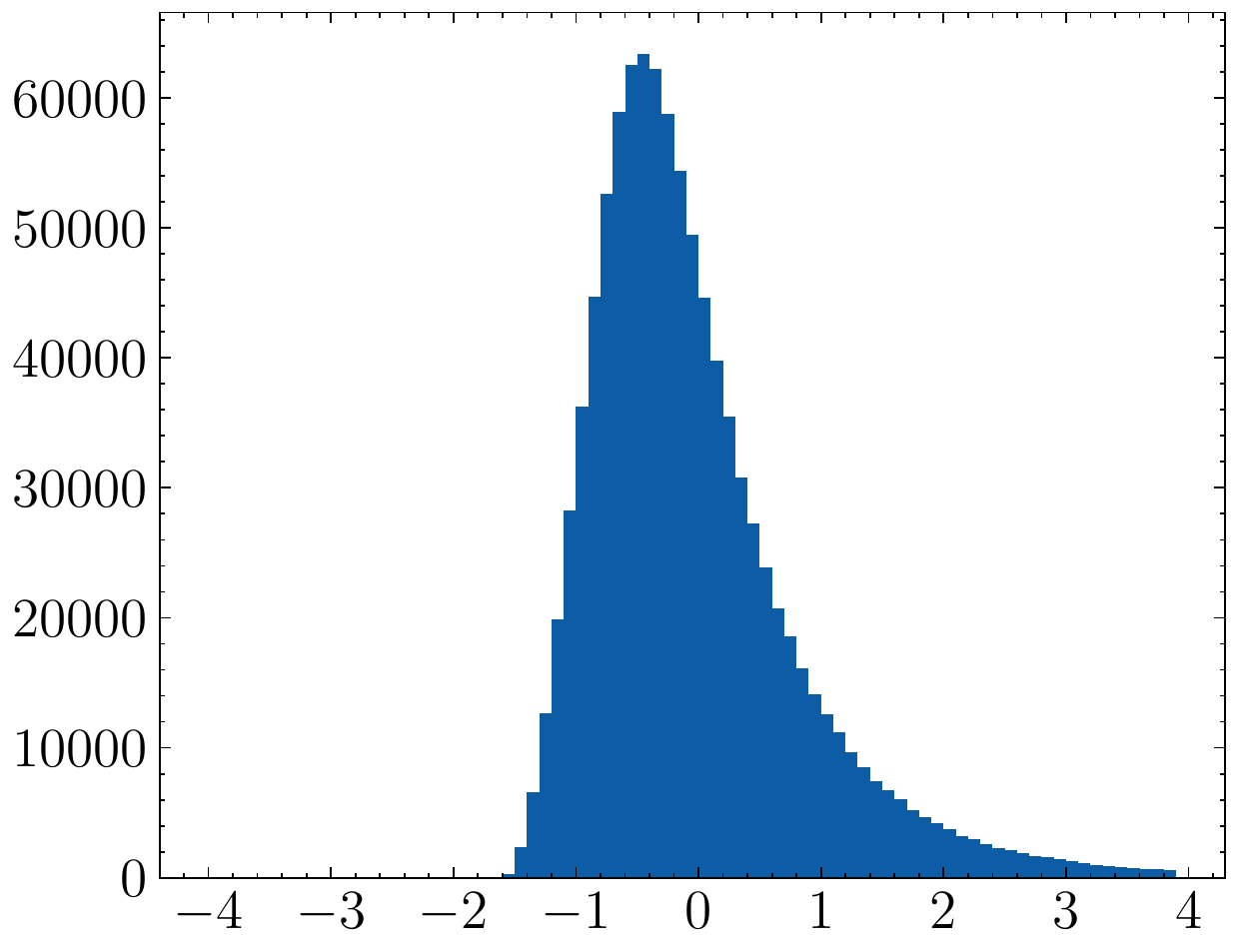}
        & \includegraphics[width=0.32\linewidth]{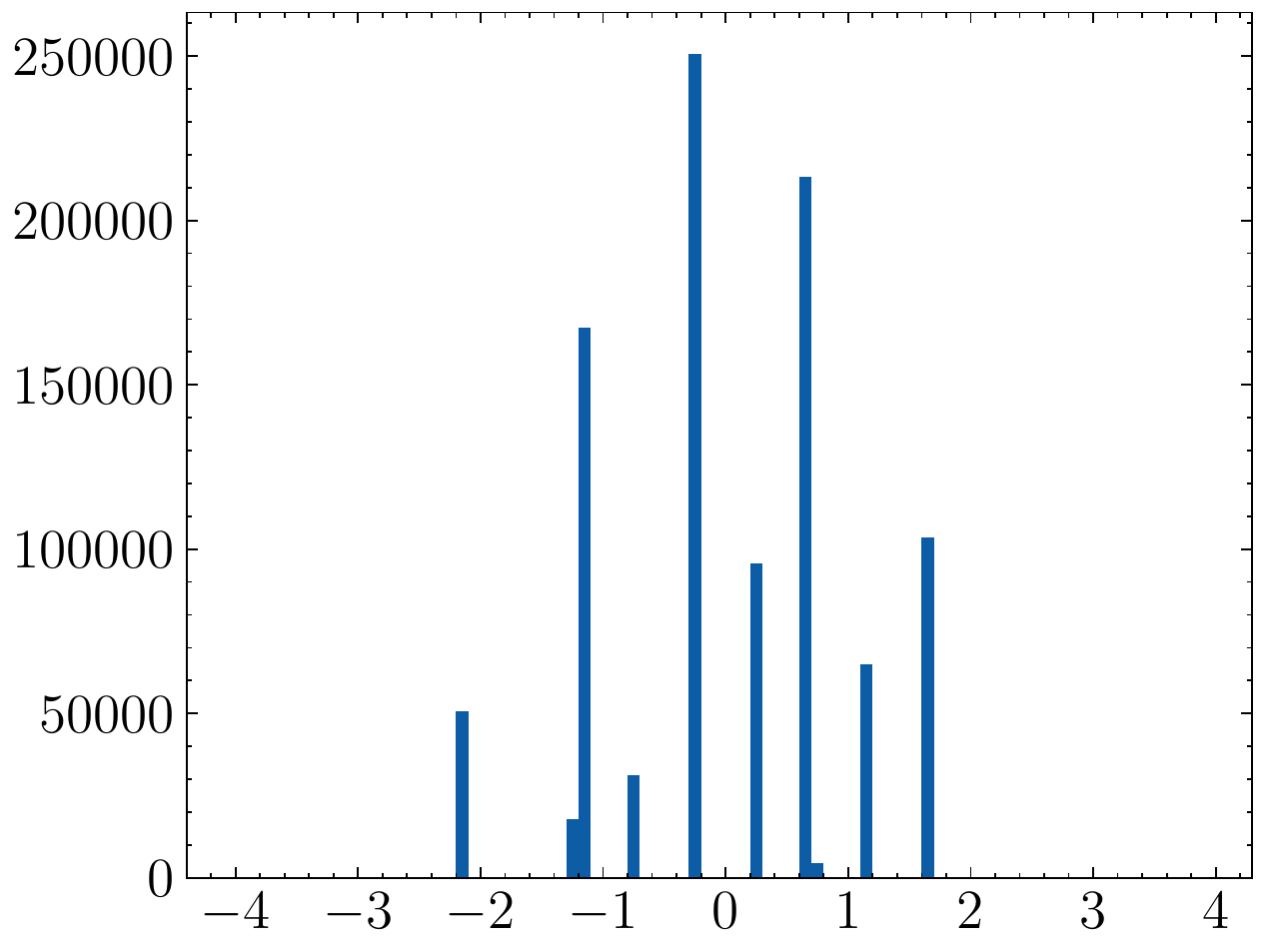}\\
        (a) & (b) & (c)
    \end{tabular}
        
    \end{center}
    \caption{Label distribution of (a) Gaussian Linear, (b) Gaussian Nonlinear Additive, and (c) Gaussian Piecewise Constant datasets. 1 million datapoints are generated for each dataset, and 120 equal sizedd bins from -6 to 6 are used for discretizing the distribution.}
\end{figure}

%% file: figures/wine_cov.tex
\begin{figure}[h]
    \centering
    \includegraphics[width=1\linewidth]{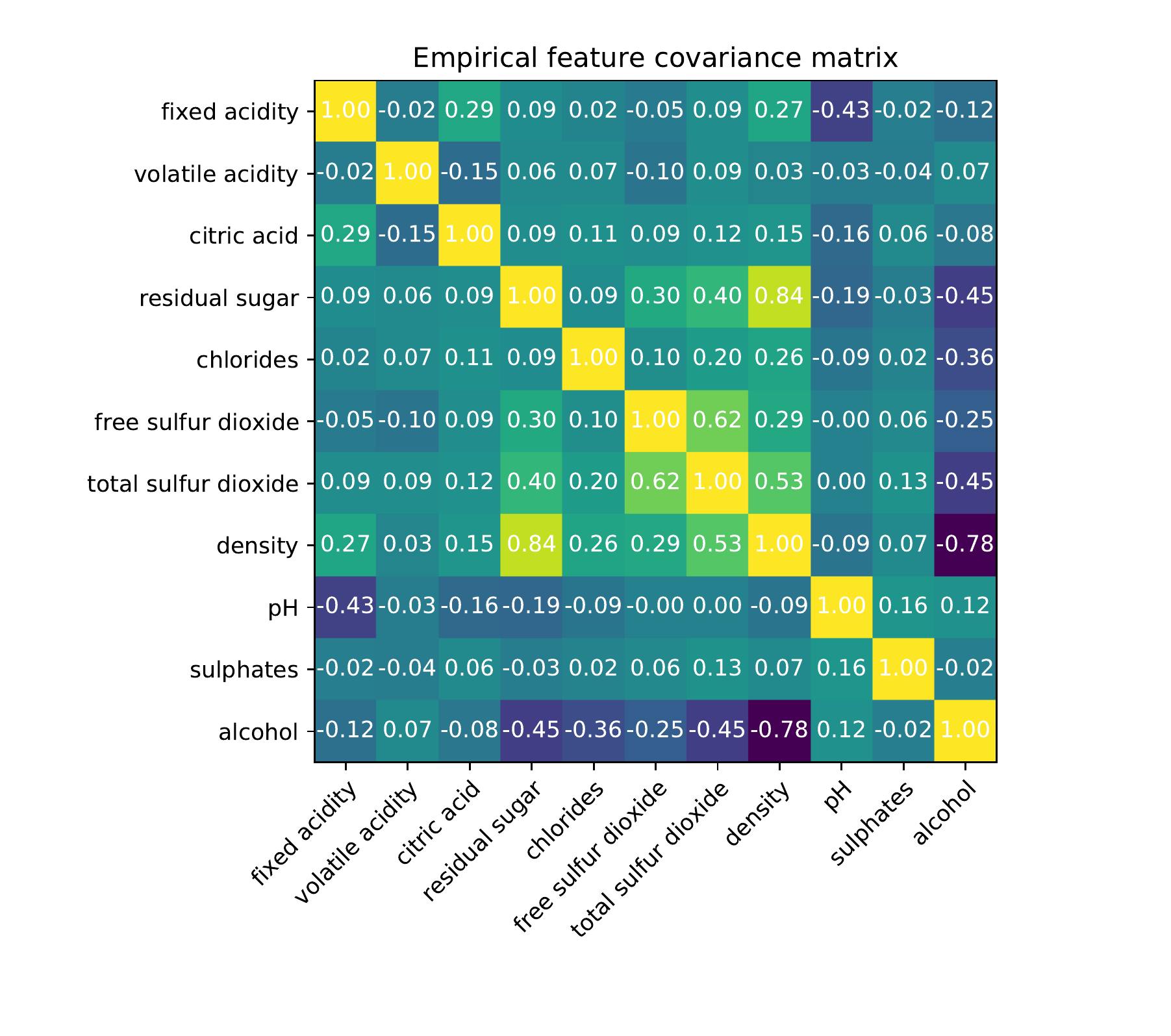}
    \caption{Empirical covariance matrix of the wine dataset. Features are normalized to have unit variance and zero mean.}
    \label{fig:wine_cov}
\end{figure}

%% file: tables/runtimes.tex
\begin{table}[h]
    \caption{Time taken in seconds by explainers to explain 100 test datapoints from the Gaussian piecewise constant dataset for a multilayer perceptron model.}
    \label{tab:runtimes}
    \vspace{2mm}
    \centering
    \begin{tabular}{c c c c c c c c}
        \hline
        & Random 
        & SHAP & SHAPR  & BF-SHAP
        & MAPLE & LIME & L2X \\ 

        \hline
        Time (in seconds) &0.00009
        &3.9&323.8&0.2&3.2&28.0&6.5\\
        
        \hline
    \end{tabular}
    \vspace{1mm}
\end{table}

%% file: tables/real_sim_mse.tex
\begin{table}[h]
    \caption{Mean squared error (MSE) between explanations for predictions of models trained on real and simulated wine dataset. Random predictions are generated from standard Gaussian distribution for every feature for each datapoint. Low MSE across ML models and explainers suggest the simulated wine dataset is a good representation of the real dataset for explainability benchmarking.}
    \label{tab:real_sim_mse}
    \vspace{2mm}
    \centering
    \begin{tabular}{c c c c c c}
        \hline
        Model 
        & SHAP %& BF-SHAP & SHAPR 
        & LIME & MAPLE & L2X 
        & Random \\ 

        \hline
        
        Linear 
        & $0.028 \pm 0.009$ %& $1.0 \pm 1.0$ & $1.0 \pm 1.0$ 
        & $0.047 \pm 0.016$ & $0.027 \pm 0.009$ & $0.0009 \pm 0.0001$
        & \multirow{3}{5.5em}{$1.988 \pm 0.001$}\\
        
        Tree
        & $0.047 \pm 0.003$ %& $1.0 \pm 1.0$ & $1.0 \pm 1.0$ 
        & $0.009 \pm 0.001$ & $0.052 \pm 0.012$ & $0.0008 \pm 0.0001$
        & \\
        
        MLP 
        & $0.028 \pm 0.003$ %& $1.0 \pm 1.0$ & $1.0 \pm 1.0$ 
        & $0.037 \pm 0.008$ & $0.040 \pm 0.002$ & $0.0008 \pm 0.0001$
        & \\
        
        \hline
    \end{tabular}
    \vspace{1mm}
\end{table}

%% file: figures/all_plots.tex
\begin{figure}[h]
    \begin{center}
        \includegraphics[width=\linewidth]{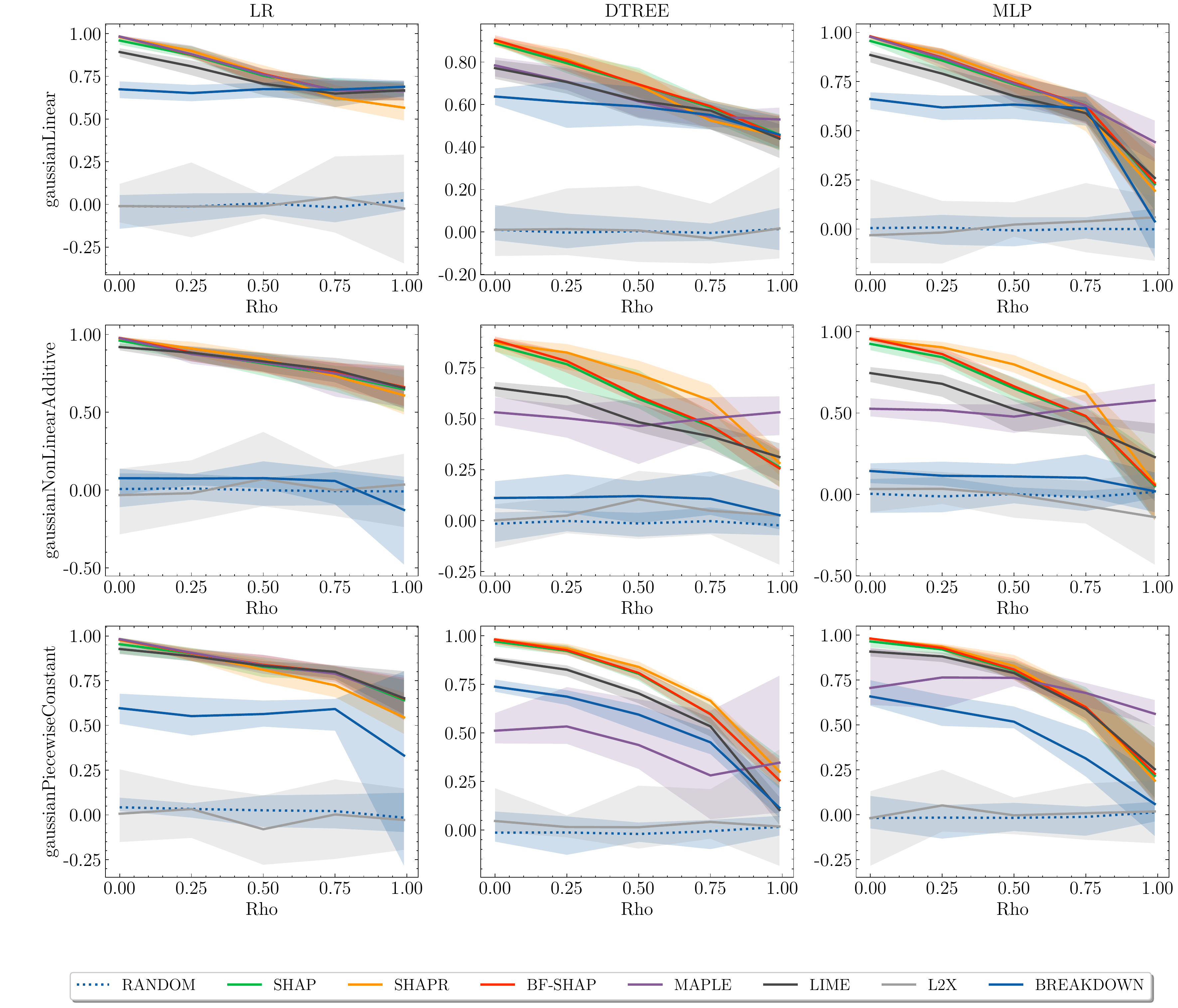}
    \end{center}
    \caption{Results of faithfulness across ML models, dataset types, and $\rho$s.}
\end{figure}

\begin{figure}[h]
    \begin{center}
        \includegraphics[width=\linewidth]{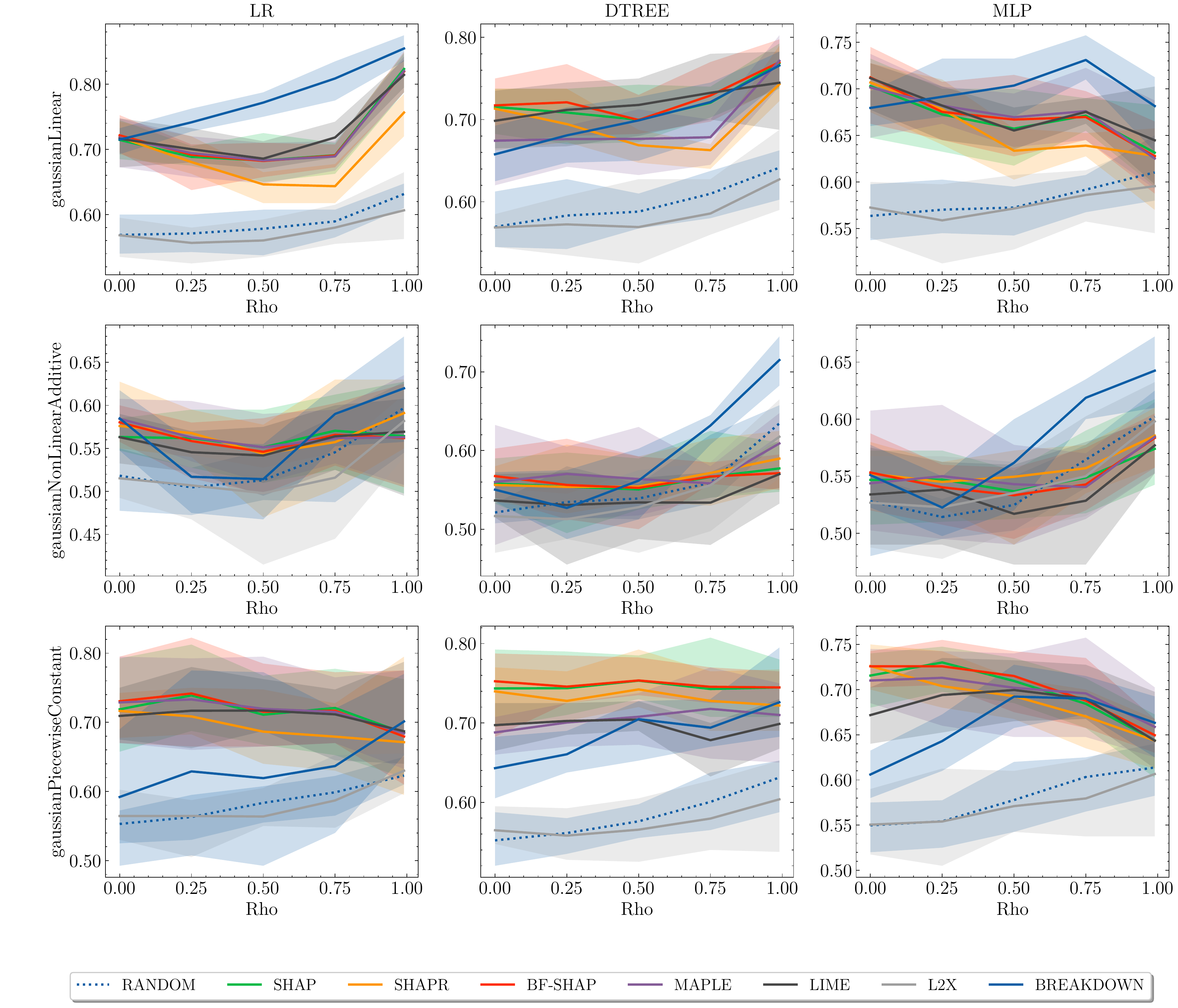}
    \end{center}
    \caption{Results of monotonicity across ML models, dataset types, and $\rho$s.}
\end{figure}

\begin{figure}[h]
    \begin{center}
        \includegraphics[width=\linewidth]{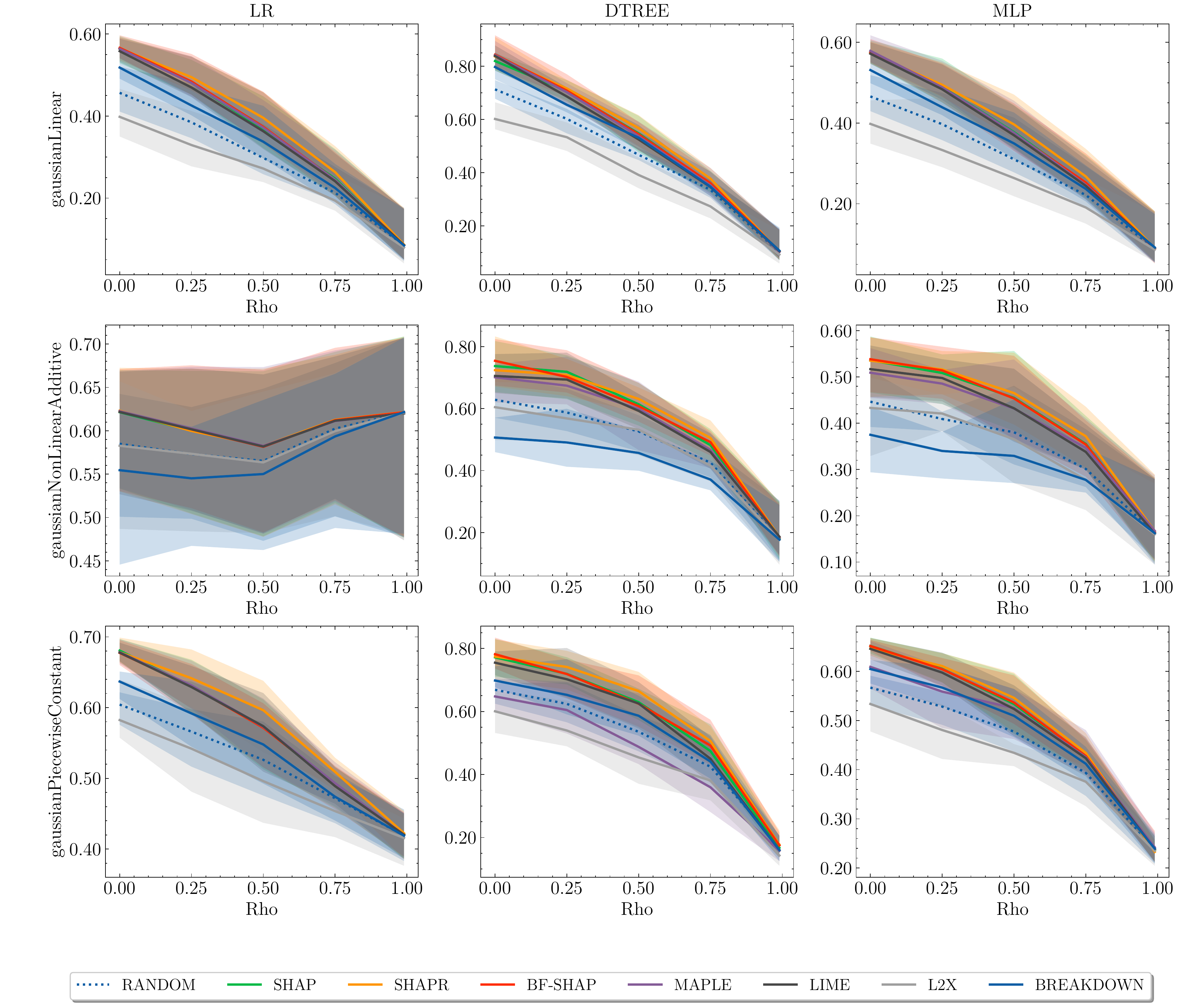}
    \end{center}
    \caption{Results of ROAR across ML models, dataset types, and $\rho$s.}
\end{figure}

\begin{figure}[h]
    \begin{center}
        \includegraphics[width=\linewidth]{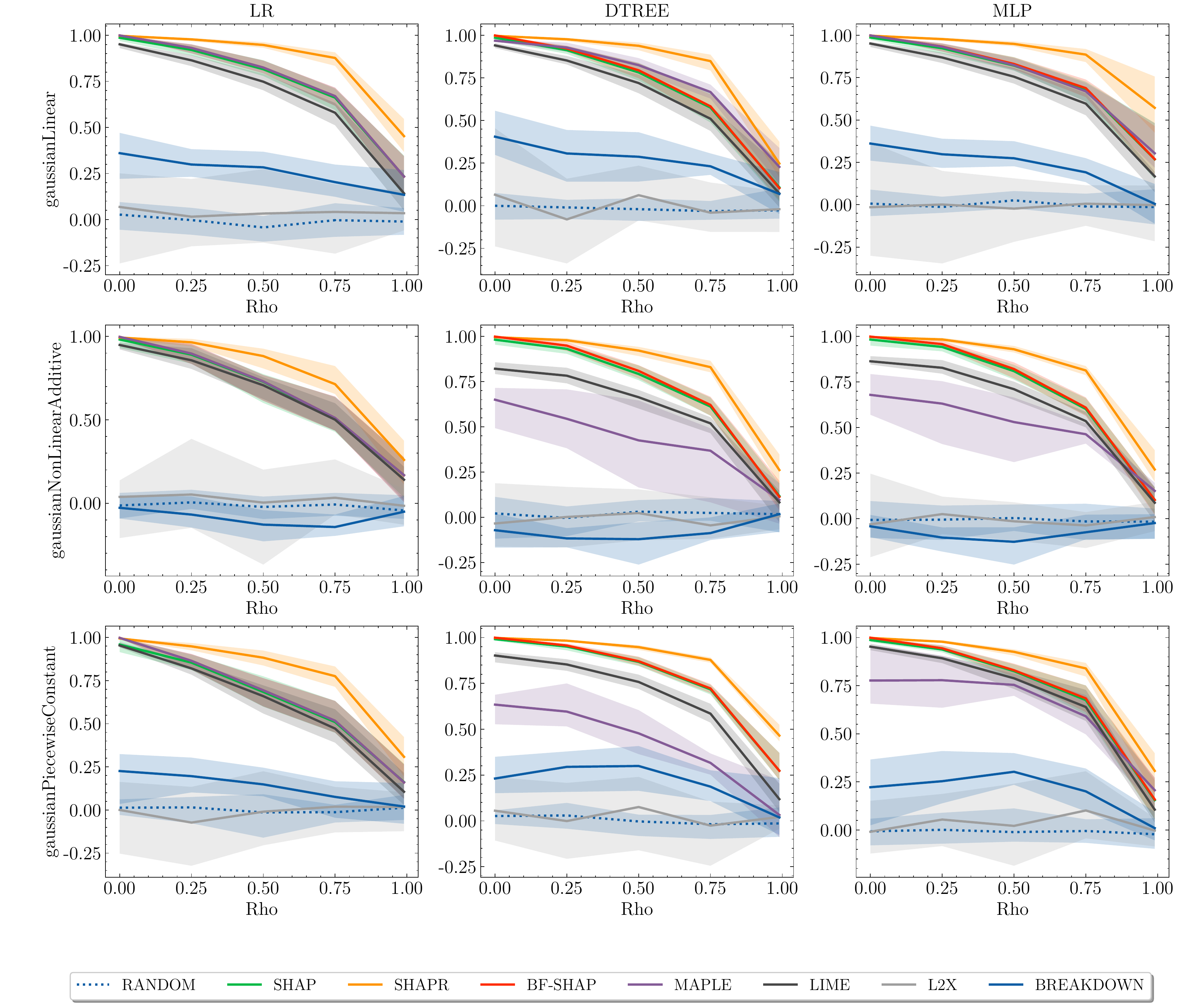}
    \end{center}
    \caption{Results of GT-Shapley across ML models, dataset types, and $\rho$s.}
\end{figure}

\begin{figure}[h]
    \begin{center}
        \includegraphics[width=\linewidth]{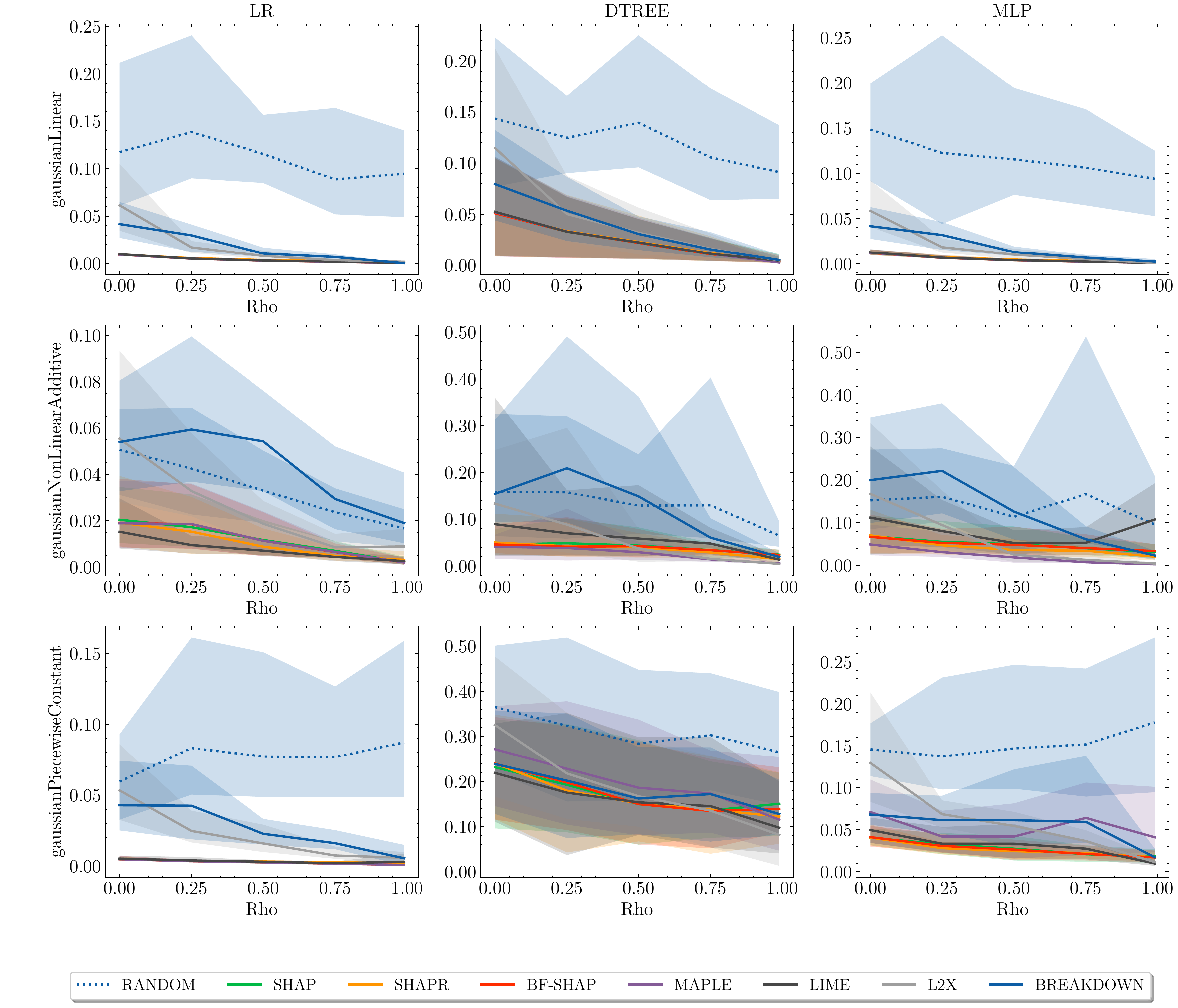}
    \end{center}
    \caption{Results of infidelity across ML models, dataset types, and $\rho$s.}
\end{figure}

%% file: main.bbl
\begin{thebibliography}{10}

\bibitem{shapr}
Kjersti Aas, Martin Jullum, and Anders L{\o}land.
\newblock Explaining individual predictions when features are dependent: More
  accurate approximations to shapley values.
\newblock {\em Artificial Intelligence}, page 103502, 2021.

\bibitem{adadi2018peeking}
Amina Adadi and Mohammed Berrada.
\newblock Peeking inside the black-box: a survey on explainable artificial
  intelligence (xai).
\newblock {\em IEEE access}, 6:52138--52160, 2018.

\bibitem{adebayo2018sanity}
Julius Adebayo, Justin Gilmer, Michael Muelly, Ian Goodfellow, Moritz Hardt,
  and Been Kim.
\newblock Sanity checks for saliency maps.
\newblock {\em arXiv preprint arXiv:1810.03292}, 2018.

\bibitem{alvarez2018towards}
David Alvarez-Melis and Tommi~S Jaakkola.
\newblock Towards robust interpretability with self-explaining neural networks.
\newblock {\em arXiv preprint arXiv:1806.07538}, 2018.

\bibitem{leaf}
Elvio Amparore, Alan Perotti, and Paolo Bajardi.
\newblock To trust or not to trust an explanation: using leaf to evaluate local
  linear xai methods.
\newblock {\em PeerJ Computer Science}, 7:e479, 2021.

\bibitem{arrieta2020explainable}
Alejandro~Barredo Arrieta, Natalia D{\'\i}az-Rodr{\'\i}guez, Javier Del~Ser,
  Adrien Bennetot, Siham Tabik, Alberto Barbado, Salvador Garc{\'\i}a, Sergio
  Gil-L{\'o}pez, Daniel Molina, Richard Benjamins, et~al.
\newblock Explainable artificial intelligence (xai): Concepts, taxonomies,
  opportunities and challenges toward responsible ai.
\newblock {\em Information Fusion}, 58:82--115, 2020.

\bibitem{arya2019one}
Vijay Arya, Rachel~KE Bellamy, Pin-Yu Chen, Amit Dhurandhar, Michael Hind,
  Samuel~C Hoffman, Stephanie Houde, Q~Vera Liao, Ronny Luss, Aleksandra
  Mojsilovi{\'c}, et~al.
\newblock One explanation does not fit all: A toolkit and taxonomy of ai
  explainability techniques.
\newblock {\em arXiv preprint arXiv:1909.03012}, 2019.

\bibitem{bansal2020sam}
Naman Bansal, Chirag Agarwal, and Anh Nguyen.
\newblock Sam: The sensitivity of attribution methods to hyperparameters.
\newblock In {\em Proceedings of the ieee/cvf conference on computer vision and
  pattern recognition}, pages 8673--8683, 2020.

\bibitem{barocas2017fairness}
Solon Barocas, Moritz Hardt, and Arvind Narayanan.
\newblock Fairness in machine learning.
\newblock {\em NIPS Tutorial}, 2017.

\bibitem{bhatt2020evaluating}
Umang Bhatt, Adrian Weller, and Jos{\'e}~MF Moura.
\newblock Evaluating and aggregating feature-based model explanations.
\newblock {\em arXiv preprint arXiv:2005.00631}, 2020.

\bibitem{bogen2018help}
Miranda Bogen and Aaron Rieke.
\newblock Help wanted: An examination of hiring algorithms, equity, and bias,
  2018.

\bibitem{chattopadhay2018grad}
Aditya Chattopadhay, Anirban Sarkar, Prantik Howlader, and Vineeth~N
  Balasubramanian.
\newblock Grad-cam++: Generalized gradient-based visual explanations for deep
  convolutional networks.
\newblock In {\em 2018 IEEE Winter Conference on Applications of Computer
  Vision (WACV)}, pages 839--847. IEEE, 2018.

\bibitem{chen2018looks}
Chaofan Chen, Oscar Li, Chaofan Tao, Alina~Jade Barnett, Jonathan Su, and
  Cynthia Rudin.
\newblock This looks like that: deep learning for interpretable image
  recognition.
\newblock {\em arXiv preprint arXiv:1806.10574}, 2018.

\bibitem{chen2020true}
Hugh Chen, Joseph~D Janizek, Scott Lundberg, and Su-In Lee.
\newblock True to the model or true to the data?
\newblock {\em arXiv preprint arXiv:2006.16234}, 2020.

\bibitem{l2x}
Jianbo Chen, Le~Song, Martin Wainwright, and Michael Jordan.
\newblock Learning to explain: An information-theoretic perspective on model
  interpretation.
\newblock In {\em International Conference on Machine Learning}, pages
  883--892. PMLR, 2018.

\bibitem{wine}
Paulo Cortez, Ant{\'o}nio Cerdeira, Fernando Almeida, Telmo Matos, and Jos{\'e}
  Reis.
\newblock Modeling wine preferences by data mining from physicochemical
  properties.
\newblock {\em Decision support systems}, 47(4):547--553, 2009.

\bibitem{forestfire}
Paulo Cortez and An{\'\i}bal de Jesus~Raimundo Morais.
\newblock A data mining approach to predict forest fires using meteorological
  data.
\newblock 2007.

\bibitem{das2020opportunities}
Arun Das and Paul Rad.
\newblock Opportunities and challenges in explainable artificial intelligence
  (xai): A survey.
\newblock {\em arXiv preprint arXiv:2006.11371}, 2020.

\bibitem{datta2016algorithmic}
Anupam Datta, Shayak Sen, and Yair Zick.
\newblock Algorithmic transparency via quantitative input influence: Theory and
  experiments with learning systems.
\newblock In {\em 2016 IEEE symposium on security and privacy (SP)}, pages
  598--617. IEEE, 2016.

\bibitem{denton2019detecting}
Emily Denton, Ben Hutchinson, Margaret Mitchell, and Timnit Gebru.
\newblock Detecting bias with generative counterfactual face attribute
  augmentation.
\newblock {\em arXiv preprint arXiv:1906.06439}, 2019.

\bibitem{deyoung2019eraser}
Jay DeYoung, Sarthak Jain, Nazneen~Fatema Rajani, Eric Lehman, Caiming Xiong,
  Richard Socher, and Byron~C Wallace.
\newblock Eraser: A benchmark to evaluate rationalized nlp models.
\newblock {\em arXiv preprint arXiv:1911.03429}, 2019.

\bibitem{dhurandhar2019model}
Amit Dhurandhar, Tejaswini Pedapati, Avinash Balakrishnan, Pin-Yu Chen,
  Karthikeyan Shanmugam, and Ruchir Puri.
\newblock Model agnostic contrastive explanations for structured data.
\newblock {\em arXiv preprint arXiv:1906.00117}, 2019.

\bibitem{dombrowski2019explanations}
Ann-Kathrin Dombrowski, Maximillian Alber, Christopher Anders, Marcel
  Ackermann, Klaus-Robert M{\"u}ller, and Pan Kessel.
\newblock Explanations can be manipulated and geometry is to blame.
\newblock {\em Advances in Neural Information Processing Systems},
  32:13589--13600, 2019.

\bibitem{dovsilovic2018explainable}
Filip~Karlo Do{\v{s}}ilovi{\'c}, Mario Br{\v{c}}i{\'c}, and Nikica Hlupi{\'c}.
\newblock Explainable artificial intelligence: A survey.
\newblock In {\em 2018 41st International convention on information and
  communication technology, electronics and microelectronics (MIPRO)}, pages
  0210--0215. IEEE, 2018.

\bibitem{fauvel2020performance}
Kevin Fauvel, V{\'e}ronique Masson, and Elisa Fromont.
\newblock A performance-explainability framework to benchmark machine learning
  methods: Application to multivariate time series classifiers.
\newblock {\em arXiv preprint arXiv:2005.14501}, 2020.

\bibitem{fu1991rule}
LiMin Fu.
\newblock Rule learning by searching on adapted nets.
\newblock In {\em AAAI}, volume~91, pages 590--595, 1991.

\bibitem{garreau2020explaining}
Damien Garreau and Ulrike Luxburg.
\newblock Explaining the explainer: A first theoretical analysis of lime.
\newblock In {\em International Conference on Artificial Intelligence and
  Statistics}, pages 1287--1296. PMLR, 2020.

\bibitem{hartley2020explaining}
Thomas Hartley, Kirill Sidorov, Christopher Willis, and David Marshall.
\newblock Explaining failure: Investigation of surprise and expectation in
  cnns.
\newblock In {\em Proceedings of the IEEE/CVF Conference on Computer Vision and
  Pattern Recognition Workshops}, pages 12--13, 2020.

\bibitem{hartley2021swag}
Thomas Hartley, Kirill Sidorov, Christopher Willis, and David Marshall.
\newblock Swag: Superpixels weighted by average gradients for explanations of
  cnns.
\newblock In {\em Proceedings of the IEEE/CVF Winter Conference on Applications
  of Computer Vision}, pages 423--432, 2021.

\bibitem{heo2019fooling}
Juyeon Heo, Sunghwan Joo, and Taesup Moon.
\newblock Fooling neural network interpretations via adversarial model
  manipulation.
\newblock {\em Advances in Neural Information Processing Systems},
  32:2925--2936, 2019.

\bibitem{roar}
Sara Hooker, Dumitru Erhan, Pieter-Jan Kindermans, and Been Kim.
\newblock A benchmark for interpretability methods in deep neural networks.
\newblock {\em arXiv preprint arXiv:1806.10758}, 2018.

\bibitem{janzing2020feature}
Dominik Janzing, Lenon Minorics, and Patrick Bl{\"o}baum.
\newblock Feature relevance quantification in explainable ai: A causal problem.
\newblock In {\em International Conference on Artificial Intelligence and
  Statistics}, pages 2907--2916. PMLR, 2020.

\bibitem{jeyakumar2020can}
Jeya~Vikranth Jeyakumar, Joseph Noor, Yu-Hsi Cheng, Luis Garcia, and Mani
  Srivastava.
\newblock How can i explain this to you? an empirical study of deep neural
  network explanation methods.
\newblock {\em Advances in Neural Information Processing Systems}, 2020.

\bibitem{lage2018human}
Isaac Lage, Andrew~Slavin Ross, Been Kim, Samuel~J Gershman, and Finale
  Doshi-Velez.
\newblock Human-in-the-loop interpretability prior.
\newblock {\em arXiv preprint arXiv:1805.11571}, 2018.

\bibitem{lakkaraju2020fool}
Himabindu Lakkaraju and Osbert Bastani.
\newblock " how do i fool you?" manipulating user trust via misleading black
  box explanations.
\newblock In {\em Proceedings of the AAAI/ACM Conference on AI, Ethics, and
  Society}, pages 79--85, 2020.

\bibitem{lakkaraju2019faithful}
Himabindu Lakkaraju, Ece Kamar, Rich Caruana, and Jure Leskovec.
\newblock Faithful and customizable explanations of black box models.
\newblock In {\em Proceedings of the 2019 AAAI/ACM Conference on AI, Ethics,
  and Society}, pages 131--138, 2019.

\bibitem{larson2016we}
Jeff Larson, Surya Mattu, Lauren Kirchner, and Julia Angwin.
\newblock How we analyzed the compas recidivism algorithm.
\newblock {\em ProPublica (5 2016)}, 9, 2016.

\bibitem{li2018deep}
Oscar Li, Hao Liu, Chaofan Chen, and Cynthia Rudin.
\newblock Deep learning for case-based reasoning through prototypes: A neural
  network that explains its predictions.
\newblock In {\em Proceedings of the AAAI Conference on Artificial
  Intelligence}, volume~32, 2018.

\bibitem{lipovetsky2001analysis}
Stan Lipovetsky and Michael Conklin.
\newblock Analysis of regression in game theory approach.
\newblock {\em Applied Stochastic Models in Business and Industry},
  17(4):319--330, 2001.

\bibitem{lundberg2017consistent}
Scott~M Lundberg and Su-In Lee.
\newblock Consistent feature attribution for tree ensembles.
\newblock {\em arXiv preprint arXiv:1706.06060}, 2017.

\bibitem{shap}
Scott~M Lundberg and Su-In Lee.
\newblock A unified approach to interpreting model predictions.
\newblock {\em Advances in Neural Information Processing Systems},
  30:4765--4774, 2017.

\bibitem{lundberg2017unified}
Scott~M Lundberg and Su-In Lee.
\newblock A unified approach to interpreting model predictions.
\newblock In {\em Advances in neural information processing systems}, pages
  4765--4774, 2017.

\bibitem{luss2019generating}
Ronny Luss, Pin-Yu Chen, Amit Dhurandhar, Prasanna Sattigeri, Yunfeng Zhang,
  Karthikeyan Shanmugam, and Chun-Chen Tu.
\newblock Generating contrastive explanations with monotonic attribute
  functions.
\newblock {\em arXiv preprint arXiv:1905.12698}, 2019.

\bibitem{meng2021mimic}
Chuizheng Meng, Loc Trinh, Nan Xu, and Yan Liu.
\newblock Mimic-if: Interpretability and fairness evaluation of deep learning
  models on mimic-iv dataset.
\newblock {\em arXiv preprint arXiv:2102.06761}, 2021.

\bibitem{molnar2019}
Christoph Molnar.
\newblock {\em Interpretable Machine Learning}.
\newblock 2019.

\bibitem{mukerjee2002multi}
Amitabha Mukerjee, Rita Biswas, Kalyanmoy Deb, and Amrit~P Mathur.
\newblock Multi--objective evolutionary algorithms for the risk--return
  trade--off in bank loan management.
\newblock {\em International Transactions in operational research}, 2002.

\bibitem{ngai2011application}
Eric~WT Ngai, Yong Hu, Yiu~Hing Wong, Yijun Chen, and Xin Sun.
\newblock The application of data mining techniques in financial fraud
  detection: A classification framework and an academic review of literature.
\newblock {\em Decision support systems}, 50(3):559--569, 2011.

\bibitem{nguyen2020quantitative}
An-phi Nguyen and Mar{\'\i}a~Rodr{\'\i}guez Mart{\'\i}nez.
\newblock On quantitative aspects of model interpretability.
\newblock {\em arXiv preprint arXiv:2007.07584}, 2020.

\bibitem{pineau2020improving}
Joelle Pineau, Philippe Vincent-Lamarre, Koustuv Sinha, Vincent Larivi{\`e}re,
  Alina Beygelzimer, Florence d'Alch{\'e} Buc, Emily Fox, and Hugo Larochelle.
\newblock Improving reproducibility in machine learning research (a report from
  the neurips 2019 reproducibility program).
\newblock {\em arXiv preprint arXiv:2003.12206}, 2020.

\bibitem{maple}
Gregory Plumb, Denali Molitor, and Ameet Talwalkar.
\newblock Model agnostic supervised local explanations.
\newblock {\em arXiv preprint arXiv:1807.02910}, 2018.

\bibitem{lime}
Marco~Tulio Ribeiro, Sameer Singh, and Carlos Guestrin.
\newblock " why should i trust you?" explaining the predictions of any
  classifier.
\newblock In {\em Proceedings of the 22nd ACM SIGKDD international conference
  on knowledge discovery and data mining}, 2016.

\bibitem{ribeiro2018anchors}
Marco~Tulio Ribeiro, Sameer Singh, and Carlos Guestrin.
\newblock Anchors: High-precision model-agnostic explanations.
\newblock In {\em Proceedings of the AAAI Conference on Artificial
  Intelligence}, volume~32, 2018.

\bibitem{ross2018improving}
Andrew Ross and Finale Doshi-Velez.
\newblock Improving the adversarial robustness and interpretability of deep
  neural networks by regularizing their input gradients.
\newblock In {\em Proceedings of the AAAI Conference on Artificial
  Intelligence}, volume~32, 2018.

\bibitem{samek2016evaluating}
Wojciech Samek, Alexander Binder, Gr{\'e}goire Montavon, Sebastian Lapuschkin,
  and Klaus-Robert M{\"u}ller.
\newblock Evaluating the visualization of what a deep neural network has
  learned.
\newblock {\em IEEE transactions on neural networks and learning systems},
  28(11):2660--2673, 2016.

\bibitem{selvaraju2017grad}
Ramprasaath~R Selvaraju, Michael Cogswell, Abhishek Das, Ramakrishna Vedantam,
  Devi Parikh, and Dhruv Batra.
\newblock Grad-cam: Visual explanations from deep networks via gradient-based
  localization.
\newblock In {\em Proceedings of the IEEE international conference on computer
  vision}, pages 618--626, 2017.

\bibitem{setiono1995understanding}
Rudy Setiono and Huan Liu.
\newblock Understanding neural networks via rule extraction.
\newblock In {\em IJCAI}, volume~1, pages 480--485. Citeseer, 1995.

\bibitem{simonyan2013deep}
Karen Simonyan, Andrea Vedaldi, and Andrew Zisserman.
\newblock Deep inside convolutional networks: Visualising image classification
  models and saliency maps.
\newblock {\em arXiv preprint arXiv:1312.6034}, 2013.

\bibitem{slack2020fooling}
Dylan Slack, Sophie Hilgard, Emily Jia, Sameer Singh, and Himabindu Lakkaraju.
\newblock Fooling lime and shap: Adversarial attacks on post hoc explanation
  methods.
\newblock In {\em Proceedings of the AAAI/ACM Conference on AI, Ethics, and
  Society}, pages 180--186, 2020.

\bibitem{srinivas2019full}
Suraj Srinivas and Fran{\c{c}}ois Fleuret.
\newblock Full-gradient representation for neural network visualization.
\newblock {\em arXiv preprint arXiv:1905.00780}, 2019.

\bibitem{breakdown}
Mateusz Staniak and Przemyslaw Biecek.
\newblock Explanations of model predictions with live and breakdown packages.
\newblock {\em arXiv preprint arXiv:1804.01955}, 2018.

\bibitem{strumbelj2010efficient}
Erik Strumbelj and Igor Kononenko.
\newblock An efficient explanation of individual classifications using game
  theory.
\newblock {\em The Journal of Machine Learning Research}, 11:1--18, 2010.

\bibitem{sundararajan2020many}
Mukund Sundararajan and Amir Najmi.
\newblock The many shapley values for model explanation.
\newblock In {\em International Conference on Machine Learning}, pages
  9269--9278. PMLR, 2020.

\bibitem{towell1993extracting}
Geoffrey~G Towell and Jude~W Shavlik.
\newblock Extracting refined rules from knowledge-based neural networks.
\newblock {\em Machine learning}, 13(1):71--101, 1993.

\bibitem{wong1985entropy}
Andrew~KC Wong and Manlai You.
\newblock Entropy and distance of random graphs with application to structural
  pattern recognition.
\newblock {\em IEEE Transactions on Pattern Analysis and Machine Intelligence},
  1985.

\bibitem{wright1921correlation}
Sewall Wright.
\newblock Correlation and causation.
\newblock {\em Journal of Agricultural Research}, 20:557--580, 1921.

\bibitem{yang2019benchmarking}
Mengjiao Yang and Been Kim.
\newblock Benchmarking attribution methods with relative feature importance.
\newblock {\em arXiv preprint arXiv:1907.09701}, 2019.

\bibitem{yeh2019fidelity}
Chih-Kuan Yeh, Cheng-Yu Hsieh, Arun Suggala, David~I Inouye, and Pradeep~K
  Ravikumar.
\newblock On the (in) fidelity and sensitivity of explanations.
\newblock {\em Advances in Neural Information Processing Systems},
  32:10967--10978, 2019.

\bibitem{zhang2018interpretable}
Quanshi Zhang, Ying~Nian Wu, and Song-Chun Zhu.
\newblock Interpretable convolutional neural networks.
\newblock In {\em Proceedings of the IEEE Conference on Computer Vision and
  Pattern Recognition}, pages 8827--8836, 2018.

\bibitem{zhang2020survey}
Yu~Zhang, Peter Ti{\v{n}}o, Ale{\v{s}} Leonardis, and Ke~Tang.
\newblock A survey on neural network interpretability.
\newblock {\em arXiv preprint arXiv:2012.14261}, 2020.

\end{thebibliography}
